\documentclass[preprint,12pt]{elsarticle}

\usepackage[a4paper, total={6.5in, 8in}]{geometry}

\usepackage{caption}
\usepackage{subcaption}
\usepackage[mathscr]{euscript}

\usepackage{amssymb}
\usepackage{amsmath}

\usepackage[dvipsnames]{xcolor}

\usepackage{mathtools}
\DeclarePairedDelimiter\ceil{\lceil}{\rceil}
\DeclarePairedDelimiter\floor{\lfloor}{\rfloor}

\usepackage{algorithm}
\usepackage{algpseudocode}

\usepackage{hyperref}

\journal{Nuclear Physics B}

\begin{document}

\begin{frontmatter}



\title{Reliable Statistical Guarantees for Conformal Predictors with Small Datasets} 


\author[1]{Miguel Sánchez-Domínguez\corref{cor1}} 
\ead{miguel.sanchez.dominguez@upm.es}
\cortext[cor1]{Corresponding author.}
\author[2]{Lucas Lacasa}
\author[1]{Javier de Vicente}
\author[1,3]{Gonzalo Rubio}
\author[1,3]{Eusebio Valero}

\affiliation[1]{organization={ETSIAE-UPM - School of Aeronautics, Universidad Politécnica de Madrid},
            addressline={Plaza Cardenal Cisneros 3}, 
            postcode={E-28040}, 
            city={Madrid},
            country={Spain}}
            
\affiliation[2]{organization={Institute for Cross-Disciplinary Physics and Complex Systems (IFISC), CSIC-UIB},
            city={Palma de Mallorca},
            country={Spain}}
            
\affiliation[3]{organization={Center for Computational Simulation, Universidad Politécnica de Madrid, Campus de
Montegancedo, Boadilla del Monte},
postcode={28660},
city={Madrid}, 
country={Spain}}

\begin{abstract}
Surrogate models --including deep neural networks and other machine learning algorithms in supervised learning-- are capable of approximating arbitrarily complex, high-dimensional input-output problems in science and engineering, but require a thorough data-agnostic uncertainty quantification analysis before these can be deployed for any safety-critical application. The standard approach for data-agnostic uncertainty quantification is to use conformal prediction (CP), a well-established framework to build uncertainty models with proven statistical guarantees that do not assume any shape for the error distribution of the surrogate model. However, since the classic statistical guarantee offered by CP is given in terms of bounds for the marginal (i.e. ensemble-averaged) coverage, for small calibration set sizes --which are frequent in realistic surrogate modelling whose prediction error needs to be quantified at different regions--, the potentially strong dispersion of the coverage distribution around its average negatively impacts the relevance of the uncertainty model's statistical guarantee, often obtaining coverages below the expected value, resulting in a less applicable framework. After providing a gentle presentation of uncertainty quantification for surrogate models for machine learning practitioners, in this paper we bridge the gap by proposing a new statistical guarantee that offers probabilistic information for the coverage of a single conformal predictor. 
We show that the proposed framework converges to the standard solution offered by CP for large calibration set sizes and, unlike the classic guarantee, still offers relevant information about the coverage of a conformal predictor for small data sizes. 
We illustrate and validate the methodology in a suite of simple and pedagogical examples, and implement an open access, user-friendly software solution that can be used alongside common conformal prediction libraries to obtain uncertainty models that fulfil the new guarantee.

\end{abstract}




\end{frontmatter}





\section{Introduction}

In many scientific and engineering disciplines, ranging from fluid dynamics or structural mechanics to astrophysics or finance,
high-fidelity numerical simulations and experiments are indispensable tools for predicting the input-output behaviour of complex systems. 
At the same time, traditional numerical simulation methods are often computationally expensive and time-consuming when aiming to realistically simulate physical processes, overall limiting their applicability in scenarios that demand rapid evaluations or extensive parametric studies. 
Across the disciplines, scientists and engineers have started to embrace the theory of machine learning (e.g. deep neural networks) \citep{James2013, Goodfellow2016} to build so-called {\it surrogate models}: 
efficient data-driven approximations of the original input-output processes. Surrogate models encompass the different ways in which e.g. the originally time-consuming numerical simulation of partial differential equations is approximated by a more efficient alternative. By leveraging available data, surrogate models offer the promise of drastically reducing computational costs while maintaining acceptable levels of predictive accuracy.

\medskip \noindent 
In the context of fluid dynamics, machine-learning based surrogate models have been considered for generic-purpose analysis \cite{Brunton2020,Vinuesa2023,Vinuesa2022} and for applications in aeroacoustics \cite{Kou2023}, geometric optimisation \cite{ramos2025transfer}, wind turbines \cite{Soler2024} or aerostructures \cite{Spiliotopoulos2024, ladron2025certifiable} among many others: see also  \cite{LeClainche2023, Brunton2021} and references therein. 
Beyond fluid dynamics, surrogate models are impacting virtually all branches of science, engineering and related industries \cite{Larranaga2018}, addressing important problems like the prediction of chaotic dynamics \cite{gilpin2024generative}, weather forecasting \cite{price2025probabilistic}, analysis of structural mechanics \cite{markou2024general}, finance \cite{dixon2020machine}, medicine \cite{rajkomar2019machine} and even scientific inquiries in particle physics \cite{radovic2018machine}, to name a few.\\
Once properly trained, the predictions generated by surrogate models offer a rapid means of approximating system behaviour compared to traditional simulation or experimental methods. However, such predictions are only useful if we can assess and bound their accuracy and attach to each prediction a given (statistically sound) uncertainty. For instance, if a surrogate model estimates that the lifetime of product will be of $10^3$ hours under specific operating conditions, this prediction alone carries little value unless we can quantify how accurate the model is, i.e. it's not the same having a prediction $10^3\pm 1$ than having a prediction $10^3\pm 500$. Likewise, the potential prediction error of such lifetime might depend on different features and, if so, it is instrumental to quantify these dependences as well.
This is precisely the role of the so-called {\it uncertainty models} (UMs), which aim to calibrate and provide {\it prediction intervals} that nuance each prediction of the surrogate model with their intrinsic uncertainty. Such prediction intervals merge the outcome of various sources of uncertainty, typically classified as aleatoric or epistemic uncertainty \cite{roy2011comprehensive}. It is only when the performance of the surrogate model is accompanied by a model quantifying its uncertainty that one can properly assess its usefulness.

\medskip \noindent 
One can build uncertainty models in a variety of ways \cite{chatfield1995model, roy2011comprehensive}, spanning from simple error estimates or more elaborate heuristics \cite{lacasa2025towards} to sophisticated neural-network error quantifiers \cite{kabir2018neural}. Indeed, one can even train a sophisticated neural network to predict the error associated to the prediction of our initial surrogate model, i.e. building an uncertainty model using the same techniques used to build surrogate models. But is this really the way to guarantee that the predictions have a (statistically rigorous) bounded uncertainty?
Observe that while using the sophisticated machinery of deep learning to build UMs seems tempting, the problem with this approach is that there is no statistical guarantee that the obtained UM will be {\it reliable}, i.e. we have no guarantee that the estimations of the UM will be representative of the actual surrogate model's error, so we might very well end up the same place as we started.\\
So the question is: once you define any `strategy' to assign an interval that quantifies the uncertainty around each prediction of our surrogate model, how can we be {\it sure} that this heuristic UM is itself reliable?
The popular framework known as \emph{Conformal Prediction} (CP) \cite{fontana2023conformal, shafer2008tutorial, balasubramanian2014conformal} precisely aims to offer such guarantees. CP is a method that allows you to build a UM from scratch that fulfils certain statistical guarantees. Furthermore, given an arbitrary, pre-defined UM (that may or may not provide any guarantee), CP also allows you to transform such UM into one that does provide a reliable statistical guarantee of its performance. UMs that are `conformalised' via conformal prediction are usually called \emph{conformal predictors}.\\
Despite their wide potential applicability, implementing and deploying this statistically-solid machinery in realistic surrogate models requires a solid background in statistical theory, that often precludes a smooth adoption of these methods by machine learning practitioners in science, engineering and industry.  This is specially crucial in safety-critical applications of surrogate modelling, which are massively emerging in recent years e.g. in the aeronautical industries \cite{lacasa2025towards, CoDANN, CoDANN2, ladron2025certifiable}.
One of the first goals of this work is thus to provide a gentle \cite{angelopoulos2021gentle} and pedagogical description of how UMs are related to surrogate models --with a specific focus on conformal prediction--, with the hope of helping surrogate model practitioners. 

\medskip \noindent 
Now, the process of building a conformal predictor from a given UM can conceptually be seen as adding a small correction to the output of the original UM, so that the corrected output achieves some guarantees. Despite their much-deserved success, the statistical guarantee that a conformal predictor provides usually loses relevance when the dataset used to build the uncertainty model is small. Unfortunately, this is often the case when e.g. one would need to estimate the error attached to a given prediction conditioned on a specific region of the input space where such prediction is made.
For instance, when predicting stress-related failure probability in aircraft design \cite{lacasa2025towards}, one would like to be able to quantify the error associated to a certain prediction of failure probability for each specific region of the aircraft. Unfortunately, often very concrete regions are only sparsely sampled within the (very costly) dataset. This, in practice, poses a strong limitation for conformal predictors that aim to quantify the location-based uncertainty of the surrogate prediction. In this work we also aim to bridge this gap by proposing a modification in the way a given UM is conformalised. Such proposed methodology constitutes the main technical contribution of this work. We will show that the method matches the guarantees offered by standard conformal prediction for large datasets and, unlike standard conformal prediction, our approach is still able to achieve relevant statistical guarantees even with small datasets, thereby guaranteeing the relevance of the UM regardless of the dataset size we have and potentially strengthening its application in various domains of science, engineering, and industry.

\medskip \noindent 
The rest of the paper is organized as follows: \autoref{sec:preliminaries} gently introduces the basics of surrogate modelling and of uncertainty models, along with some notation and illustrative examples. We also introduce the polysemic notion of {\it coverage} as the main metric used to assess the quality and performance of an uncertainty model. \autoref{sec:cov} provides a more in-depth tutorial of the nuances of the coverage, coverage distribution, nominal coverage and marginal coverage. After these, in \autoref{sec:conf_pred} we describe the standard conformal prediction framework, the current statistical guarantees it offers and flag their limitation for small datasets. The main novel technical contribution of this work is then presented in \autoref{sec:new_guarantee}, where we introduce a new statistical guarantee for conformal predictors that remains relevant in small datasets, and propose a method to build conformal predictors that fulfil this new guarantee. We validate the methodology and illustrate it in a suite of examples, and also implement the method in an open access Python software \cite{repo}. Finally, in \autoref{sec:discussion} we offer some concluding remarks and discuss possible applications of this work.

\section{Preliminaries: surrogate models and uncertainty models}
\label{sec:preliminaries}

In this section we cover the basics of surrogate modelling and uncertainty models, laying out the concepts and notation we will use in the following sections of this work.

\subsection{Surrogate Modelling}
Surrogate models are defined in this work as a non-linear mapping $\mathscr{F}$ that takes as input a collection (vector) of features $\mathbf{X} \in S_X$ and outputs a collection (vector) of predictions $\hat{\mathbf{Y}} \in S_Y$. Usually, $S_X \subset \mathbb{R}^n$, and $S_Y \subset \mathbb{R}^m$ for regressors (or $S_Y \subset \left[0,1\right]^m$ for soft-classifiers), where $n$ is the number of input features the surrogate model takes, and $m$ is the number of output variables (or classes if we are using a soft-classifier). Formally:

\begin{equation}
    \mathscr{F} : S_X \to S_Y, \hat{\mathbf{Y}} = \mathscr{F}(\mathbf{X})
\end{equation}

The mapping $\mathscr{F}$ aims to {\it approximate} the ground-truth input-output relationship. For example, if we are solving a multi-class animal image classification problem, the feature vector $\mathbf{X}$ encapsulates information of the image we want to classify, the outputs $\hat{\mathbf{Y}}$ are the estimated probability for the image to pertain to each considered class (dog, cat, wolf, giraffe, etc.), and $\mathscr{F}$ approximates the decision function that, given an image, assigns its ground-truth class $\mathbf{Y}=(0,0,\dots,0,1,0,0,\dots)$ (i.e. the probability is one for the true class and zero otherwise). As another example, if we are trying to determine the state of a pendulum (angle $\theta$ and angular velocity $\dot\theta$), the inputs of our model could be the initial conditions of the pendulum and the time for which we want our prediction, the outputs would be the estimated state of the pendulum at that time, and the function we are trying to approximate would be the set of equations that describe its motion. Finally, in a fluid-dynamic application, we could aim to predict the pressure coefficient at a given region of the aircraft's surface, and the feature vector $\bf X$ could include the location of this region, the geometry of the aircraft's surface and the physical properties (Reynolds and Mach numbers, angle of attack), whereas $\mathscr{F}$ aims to approximate the resulting input-output relationship which results of integrating the Navier-Stokes equations of the fluid flow field over the specific geometry and the subsequent derivation of the pressure coefficient.

\medskip \noindent
In the context of machine learning, the mapping $\mathscr{F}$ is \emph{learned} from a set of training data or \emph{training set} $\texttt{Train} = \left\{(\mathbf{X,\mathbf{Y})}_i\right\}_{i=1}^{N_\text{train}}$ by finding the parametric representation $\mathscr{F}$ that better matches the ground truth $\mathbf{Y}_i$ and the prediction $\hat{\mathbf{Y}}_i$ for all $\mathbf{X}$ in the set.
The universal approximation properties of neural networks \cite{hornik1990universal, hornik1991approximation} suggest that, no matter how complex the input-output relationship encoded in the training data, one can always build good approximations $\mathscr{F}$ by stacking sufficiently many and wide enough layers of neurons.
However, the universal function approximation theorems do not guarantee perfect interpolation in unseen data, and thus the generalisation performance will inherently suffer from prediction errors. Furthermore, training data is by construction finite, and thus any learned representation that uses finite data will only be an approximation to the true input-output relationship. Finally, potential errors present in the training data can also bias such learned representation.
Accordingly, once the best mapping $\mathscr{F}$ is learned, we need to evaluate its (generalisation) performance in a set of previously unseen data, usually called \emph{test set}: $\texttt{Test}=\left\{(\mathbf{X,\mathbf{Y}})_i\right\}_{i=1}^{N_\text{test}}$.

\medskip \noindent
A thorough quantification of the prediction errors in unseen data is especially important in safety-critical domains such as aerospace engineering, where the deployment of surrogate models require a rigorous process of certification that goes beyond global error characterisation \cite{lacasa2025towards}. 
In this context, the surrogate model's outputs should not be just point predictions, but {\it prediction sets} that quantify the goodness of the approximation --and the uncertainty around it-- the surrogate model is doing. These prediction sets (which, in the case of regression, are called {\it prediction intervals}) offer a collection of possible outputs (somewhat centered at $\hat{\mathbf{Y}}_i$) where we expect the ground truth $\mathbf{Y}_i$ to be. Following the previous examples, in the case of the classification of images of animals, if we are presented with the image of a big, grey canine, one prediction set could be $\left\{\text{wolf, dog}\right\}$, and in the case of the pendulum a prediction interval could be $\theta \in \left[-0.1, 0.2\right]\text{rad}$, $\dot\theta \in \left[0.1, 0.15\right]\text{rad/s}$.\\
The requirement to build the aforementioned prediction sets creates the need for a method to quantify and estimate the error of the surrogate model. This is where uncertainty models (UMs) play a crucial role.

\subsection{Uncertainty Models}
\label{sec:ums}

Uncertainty models (UMs) provide a principled framework for quantifying the uncertainty of surrogate model predictions and their expected discrepancy with respect to ground truth behaviours. For instance, these UMs allow us to faithfully estimate the previously described prediction sets. 
As for surrogate models, uncertainty models are also learned from a set of data, albeit usually \emph{different} from the one used to train the surrogate $\mathscr{F}$, and often called the \emph{calibration set}: $\texttt{Cal}=\left\{(\mathbf{X,\mathbf{Y})}_i\right\}_{i=1}^{N_\text{cal}}$ (accordingly, the total labelled dataset is often partitioned (or split) into disjoint training, calibration and test sets, and such partition needs to fulfil certain statistical properties \cite{goodfellow2016deep}). Once built on the calibration set, the UMs are evaluated on a test set that is usually the same set used to evaluate the surrogate $\mathscr{F}$.\\
Formally, an uncertainty model can be seen as a function $\mathscr{U}$ that takes data $\{\bf X, \bf Y\}$ from the calibration set along with the predictions $\hat{\bf Y}=\mathscr{F}(\bf X)$ and outputs an uncertainty metric $\bf U$ (for instance, the absolute error in regression problems), such that ${\bf U}=\mathscr{U}(\bf X, Y,\hat{Y})$.



The estimated uncertainty metric $\mathbf{U}$ is then used to build a prediction set $\mathscr{S}$, and the whole process of associating $\mathscr{S}$ to each prediction $\mathscr{F}({\bf X})\to \mathscr{S}$ defines the UM. For illustration, in a regression problem, one could simply define $\mathbf{U}$ as an estimate of the absolute error in the calibration set $\mathbf{U} = \langle |\hat{\mathbf{Y}}_i - \mathbf{Y}_i| \rangle_{i\in \texttt{Cal}}$ and the prediction interval associated with this uncertainty measure could simply be $\mathscr{S} = \left[\hat{\mathbf{Y}}-\mathbf{U}, \hat{\mathbf{Y}}+\mathbf{U}\right]$. If the chosen uncertainty metric were exactly estimated, we could guarantee that the estimated prediction would contain the true value $\mathbf{Y}$ for a certain confidence level (i.e. we could guarantee $\mathbf{Y}\in\mathscr{S}$ with certain confidence). However, since the uncertainty model only provides an estimation of the uncertainty metric and the calibration set is finite, the prediction sets built from it cannot guarantee when and how many of the true outputs are contained in each $\mathscr{S}$. 
Note that, in the rest of this work, for the sake of illustration we will be building prediction intervals of scalar uncertainty models (i.e. ${\bf U} \in \mathbb{R}$ and $S_Y \subset \mathbb{R}$) associated to regressors $\mathscr{F}$ that take values on a vector ${\bf X} \in \mathbb{R}^n$ and output a scalar ${\bf Y}\in \mathbb{R}$, but for notation consistency we will be maintaining the bold font for $\bf U$, $\bf Y$ and $\hat{\bf Y}$.

\medskip \noindent 
How to assess the quality of a UM? The proportion of cases in which the estimated prediction set $\mathscr{S}$ contains the true value $\bf Y$ for all $\bf Y$ in the test set is called the {\it coverage} $C$ of the UM. Now, since the calibration set is finite, $C$ is itself a random variable exposed to finite size fluctuations which are dependent on the specific calibration set sampling used to build the prediction sets (i.e. the coverage of the UM will be different for each sample used as calibration set). The important notion of the confidence of the UM or more popularly the \emph{marginal coverage} of the UM emerges when one averages the calibration set-dependent $C$ over an ensemble of different calibration set samplings\footnote{Hence the word `marginal', as we are marginalizing (ensemble-averaging) the coverage --which depends on the calibration set sampling-- over the ensemble of compatible calibration set samplings.}. Formally, the marginal coverage of the UM is the probability that any given point in the test set is within the predicted set: $\mathbb{P}({{\bf Y} \in \mathscr{S}}) = \mathbb{E}(C)$, where the expectation value is taken over an hypothetical ensemble of calibration sets.


\medskip \noindent 
Often, UMs are built by attempting to prescribe coverages sufficiently close to one (although this of course depends on the specific application). Note, however, that coverages extremely close to one come at the price of losing {\it informativeness}: as enforcing very large coverages usually enlarges the size of the prediction sets and in the limit one can end up with essentially useless prediction sets: think how useless is a prediction like $0.7 \pm 100$.\\
Then, and as the reader will suspect, the \emph{desired} marginal coverage the user aims to prescribe (often called the {\it nominal} coverage, $C^{\text{nom}}$) when building a UM happens to differ in practice with the \emph{actual} marginal coverage, again because prediction sets are built on the calibration set and thus suffer from finite size effects.
These deviations are explained in depth in \autoref{sec:cov}. In any case, the desired marginal coverage $C^{\text{nom}}$ is usually specified by the user as a number \emph{close} to one (typically 0.9, 0.95, 0.99, etc.). Then, the deviation from this nominal prescription of the actual marginal coverage that is effectively provided by the UM needs to be bounded: these bounds are called a {\it statistical guarantee}. For example, in \autoref{sec:simple_um} we will describe a very simple UM that provides the guarantee

\begin{equation}
    \frac{-1}{N_\text{cal}+1} \leq \mathbb{P}(\mathbf{Y}\in\mathscr{S}) - C^\text{nom} \leq \frac{1}{N_\text{cal}+1}
    \label{eqn:guarantee_simple}
\end{equation}

\noindent where $\mathbb{P}(\mathbf{Y}\in\mathscr{S})$ is the marginal coverage of the UM and $C^\text{nom}$ represents the desired marginal coverage specified a priori by the user. This guarantee means that the difference between the desired marginal coverage $C^{\text{nom}}$ and the actual marginal coverage is bounded in an interval whose bounds depend on the calibration set size $N_\text{cal}$. In a nutshell, this guarantee tells us something very intuitive: the bigger the calibration set, the closer the actual marginal coverage will be to the desired one.\\
To summarize, a valid UM needs not only to be {\it informative}, i.e. the size of the prediction sets $\mathscr{S}$ need to be small enough so that the UM actually provides non-trivial information of the uncertainty around each prediction of the surrogate model (note sometimes informativeness is also called {\it efficiency}), but also needs to be {\it reliable}: the UM must fulfil some kind of statistical guarantee. Traditionally, this guarantee is provided in terms of its marginal coverage, so that it is as close as possible to the desired marginal coverage specified by the user. Validation of this requirement is usually done by direct inspection of the {coverage} of the UM measured in a test set. Since the guarantee is expressed in terms of marginalised quantities over an ensemble of calibration sets (i.e. the guarantee provides information about the expected coverage and not about the actual coverage a specific UM will get), we advance that the relevance of these statistical guarantees will be challenged when the size of the calibration set sampling is small, as in this case deviations from the average behaviour can be significant.



\begin{figure}[!h]
    \centering
    \begin{subfigure}[b]{0.3\textwidth}
        \centering
        \includegraphics[width=0.7\textwidth]{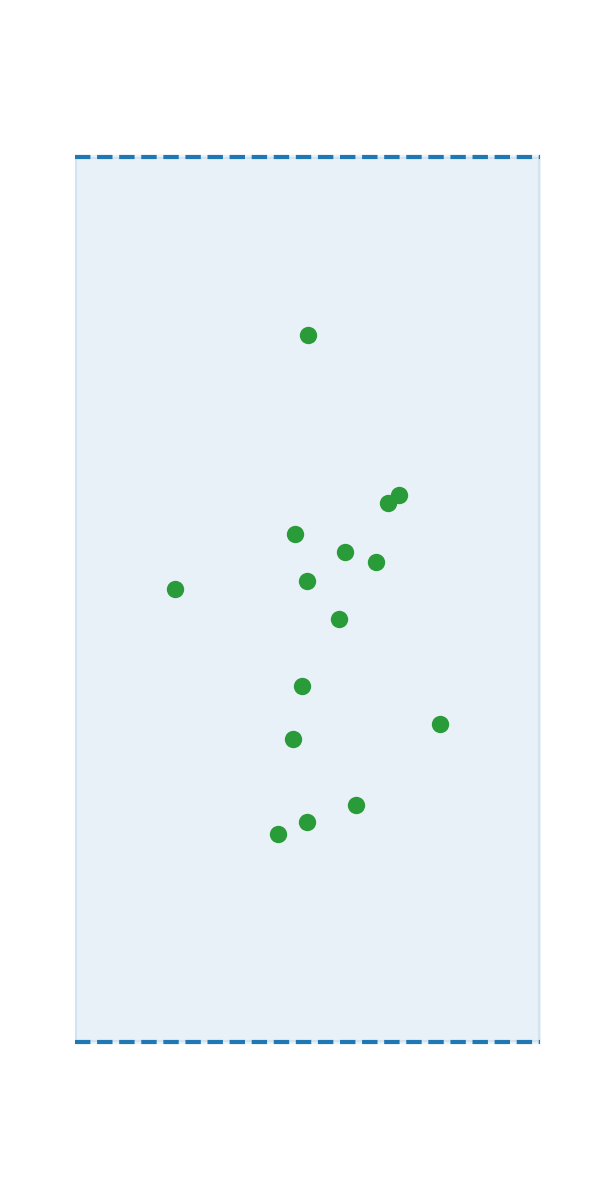}
        \caption{The prediction interval is reliable, but provides no information as it is too wide.}
        \label{fig:pred_interval_demo_rel}
    \end{subfigure}
    \hfill
    \begin{subfigure}[b]{0.3\textwidth}
        \centering
        \includegraphics[width=0.7\textwidth]{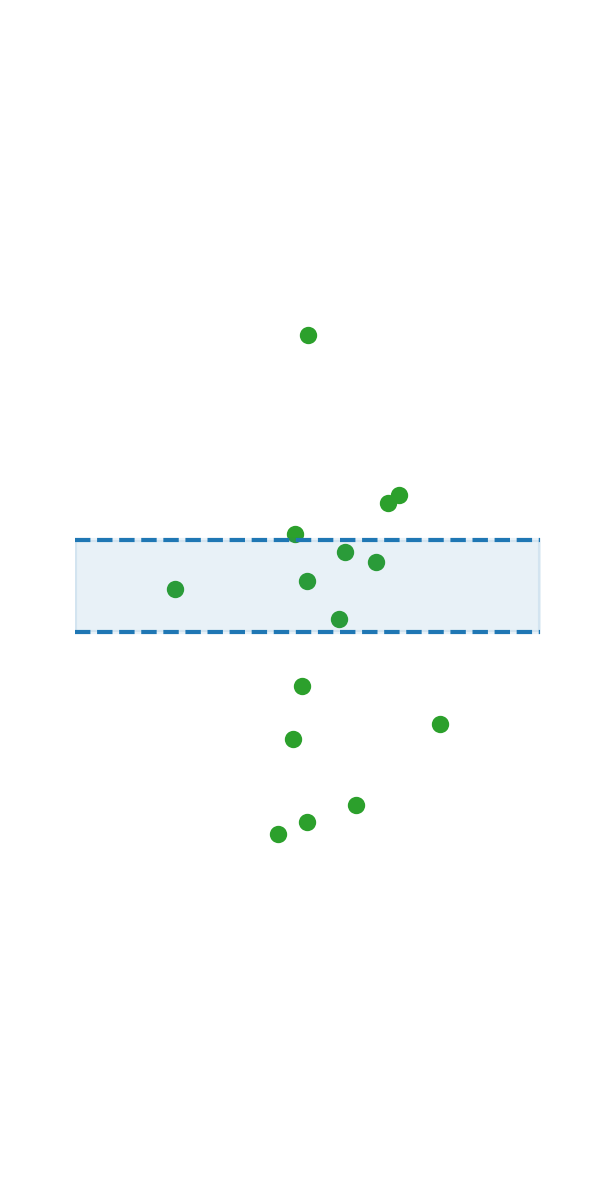}
        \caption{The prediction interval is informative, but not reliable as a considerable proportion of examples lie outside of it.}
        \label{fig:pred_interval_demo_inf}
    \end{subfigure}
    \hfill
    \begin{subfigure}[b]{0.3\textwidth}
        \centering
        \includegraphics[width=0.7\textwidth]{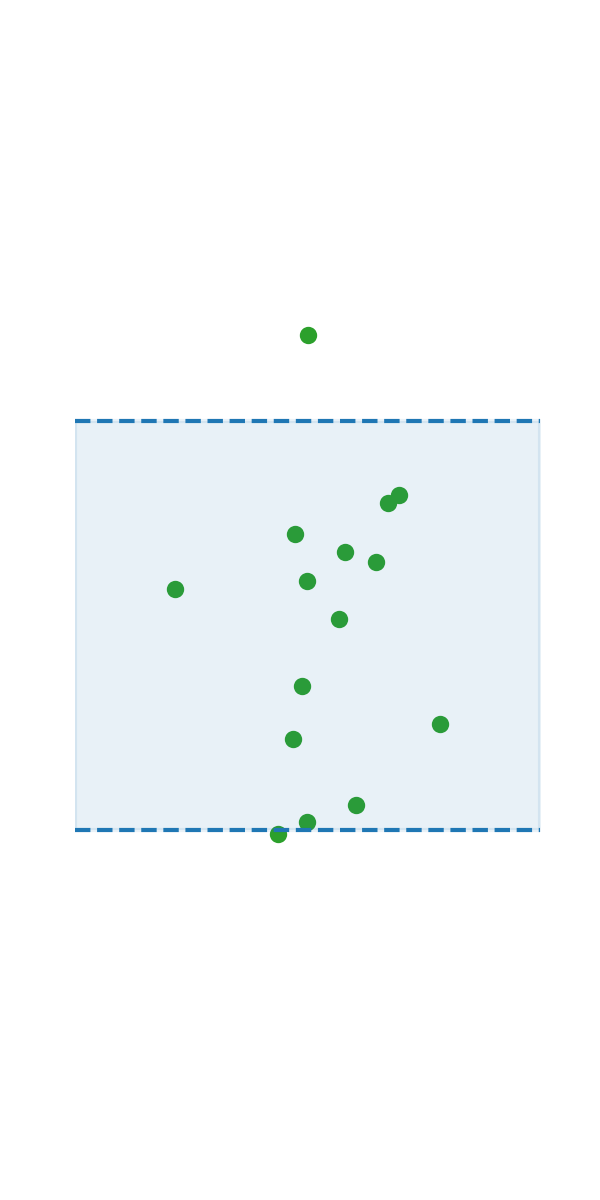}
        \caption{The prediction interval has a good compromise between being informative and reliable.}
        \label{fig:pred_interval_demo_rel_inf}
    \end{subfigure}
    \caption{Schematic of the meaning of a UM being reliable and informative. Prediction intervals are represented by blue shaded areas, and test examples by green dots.}
    \label{fig:pred_interval_demo}
\end{figure}

\medskip \noindent 
To further illustrate what these two properties (being reliable and being informative) mean, in \autoref{fig:pred_interval_demo} we show three examples of prediction intervals, one very reliable but non-informative as the interval is too wide, one very informative but not reliable as a big proportion of test points fall outside the prediction interval, and a last one that has a good compromise between reliability and informativeness as it manages to capture a considerable proportion of test examples inside the prediction interval while  keeping the width of the interval reasonably small.

\subsection{A simple uncertainty model}
\label{sec:simple_um}

To conclude this brief introduction of concepts, we illustrate how to build a very simple uncertainty model.
In later sections we will provide a detailed explanation of how we can build better uncertainty models using the conformal prediction framework, and how can we improve the statistical guarantees that these models provide when trained on small datasets, the latter being the main technical contribution of this work.\\
Imagine that we have trained a surrogate regressor $\mathscr{F}$ on some training data $\texttt{Train}=\left\{(\mathbf{X,\mathbf{Y})}_i\right\}_{i=1}^{N_\text{train}}$, and that we have a calibration set of unseen data $\texttt{Cal} = \left\{(\mathbf{X,\mathbf{Y})}_i\right\}_{i=1}^{N_\text{cal}}$ that we want to use to build a UM with a desired marginal coverage $C^\text{nom}=0.9$. A very simple approach would involve:

\begin{enumerate}
    \item Computing the absolute error of the surrogate model $\mathscr{F}$ on each example of the calibration set. This would give us an empirical approximation of the (absolute) error distribution of the surrogate $\mathscr{F}$.
    \item Computing the sample quantile $\hat{Q}_q$ at level $q=0.9$ of the absolute error distribution, which will act as the estimated uncertainty $\mathbf{U}$ of our model. Note that setting a fixed $\mathbf{U}$ which is independent of $\bf X$ and $\bf Y$ implicitly assumes that the error is uniformly distributed over the domain of $\mathscr{F}$.
    \item Once we have $\mathbf{U} = \hat{Q}_q$, the prediction interval would simply be $\mathscr{S}=\left[\hat{\mathbf{Y}} - \hat{Q}_q, \hat{\mathbf{Y}} + \hat{Q}_q\right]$.
\end{enumerate}

Although simple, the previous steps summarize the workflow common to all uncertainty models we will study in this work:

\begin{enumerate}
    \item Choose an error metric (in the previous example, the absolute error) and evaluate it on the calibration set, obtaining an empirical distribution of such metric.
    \item Obtain a sample quantile on the previously computed empirical distribution. The specific quantile level we use and how it relates to the desired marginal coverage will be a central part of this work later on, in sections \ref{sec:conf_pred} and \ref{sec:new_guarantee}.
    \item Use the computed sample quantile to define the prediction set.
\end{enumerate}

As stated before, at this point we will not be able to guarantee that the actual marginal coverage provided by the UM is equal to $C^\text{nom}$, but we shall be able to bound its deviation. Specifically, this simple approach returns prediction intervals that fulfil the statistical guarantee given by \autoref{eqn:guarantee_simple} (the reader can find how this guarantee is derived in \autoref{sec:simple_um_guarantee_dem}).

\section{In-depth discussion on the polysemic notion of coverage}
\label{sec:cov}

As seen in \autoref{sec:preliminaries}, the concept of coverage and marginal coverage are central pieces for defining and evaluating uncertainty models. In this section, we need to depict a more in-depth presentation on how both are defined and how we can characterise them.

\subsection{Sample quantile estimators}
\label{sec:sample_quantile_estimators}

In the example presented in \autoref{sec:simple_um} we enumerated the key steps to build a UM, being one of the steps the estimation of a sample quantile. In what follows, we will consider that the sample quantiles are estimated using order statistics, this being one of the simplest and most used approaches within uncertainty quantification and conformal prediction
\footnote{The properties of the UM coverage are highly dependant on the choice of the quantile estimator. Since the conformal prediction framework assumes that the quantiles are computed with order statistics, we will use this method.}. With this in mind, given a generic set (a sample) of $N$ points $\left\{s_1, s_2, ...,s_N\right\}$, the sample quantile at level $q$ is defined as
\begin{gather}
    \label{eqn:samp_quantile}
  \hat{Q}_q = s_{m:N} \qquad 0 < q < 1\\
  m = \ceil{N\cdot q} 
\end{gather}
\noindent where $s_{m:N}$ is the $m$-th order statistic of the sample (i.e., the $m$-th smallest value), and $\ceil{\cdot}$ is the ceiling function. This definition of sample quantile is equivalent to the definition 1 from \cite{hyndman1996sample}, and is implemented in most statistical packages. For instance, for the sample $\left\{s_i\right\}_{i=1}^N=\left\{1, 3, 3, 4, 5, 7, 8, 9, 9, 12, 15\right\}$ the sample quantile at level $q=0.9$ is simply $\hat{Q}_{0.9}=s_{10:11}=12$ since $m=\ceil{N\cdot q}=\ceil{11 \cdot 0.9}=10$.



\subsection{Coverage distribution}

As presented in \autoref{sec:ums}, once the UM is built using the (finite) calibration set, the actual coverage of a UM is the proportion of test examples that fall inside the estimated prediction set in the (unseen) test set. Now, the fact that the calibration set is finite induces some finite size sampling effects. In particular, $\hat{Q}_q$, the estimated sample quantile at level $q$ of the error metric of choice evaluated in the calibration set (which here we generically call $s$ in allusion to the preceding subsection), generally differs from the true quantile ${Q}_q$ of the underlying error distribution. This true quantile ${Q}_q$, computed at quantile level $q$ in the true (yet unobservable) error distribution, would only be recovered in the limit of infinitely large calibration sets. The deviation between $\hat{Q}_q$ and $Q_q$ can be effectively interpreted as if we were computing a different quantile level $\hat{q}$ from the true distribution. More concretely, labelling $F_s$ as the (true) cumulative distribution function of the surrogate model's error metric of choice, the quantile's level $q$ of the true error distribution is defined as $q=F_s({Q}_q)=\mathbb{P}(s\leq {Q}_q)$. However since $\hat{Q}_q$ deviates from ${Q}_q$, this mismatch can be rewritten as effectively computing a different level $\hat{q}$ from the true distribution, where
\begin{equation}
    \hat{q} = F_s(\hat{Q}_q) \neq q = F_s(Q_q)
\end{equation}

\noindent We have illustrated this effect in Figure ~\ref{fig:samp_vs_true_percent}. Once the quantile estimation $\hat{Q}_q$ is computed, we can follow the steps to continue building the UM and then measure its coverage on our test set. Assume for now that the test set is itself infinite, so that the distribution of the surrogate’s error over the test set matches the ground-truth error distribution assumed for the calibration set. Since the estimated quantile $\hat{Q}_q$ corresponds to a probability level $\hat{q} \neq q$, the resulting coverage will be $C = \hat{q}$. If we were to repeat this procedure with a different sampling of the calibration set, we would obtain a different $\hat{Q}_q$ and therefore a different coverage. This observation outlines the random nature of the coverage, which, as previously discussed, will depend on the sampling done to obtain the calibration set.

\begin{figure}[!h]
    \centering
    \includegraphics[width=0.8\textwidth]{figures/sample_vs_true_percentile.png}
    \caption{Comparison between the true distribution quantile $Q_q$ and the sample quantile $\hat{Q}_q$. Both quantiles have been computed at level $q$ but, due to the finite size of the sample, the sample quantile $\hat{Q}_q$ reaches a level $\hat{q} \neq q$.}
    \label{fig:samp_vs_true_percent}
\end{figure}

In \ref{app:cov_dist} we show that, according to this source of variability, the coverage $C$ is a random variable which is Beta-distributed:

\begin{equation}
    C \sim \text{Beta}(m, N_\text{cal}-m+1)
\end{equation}

\noindent with $m = \ceil{N_\text{cal}\cdot q}$. With this result, we can easily see that, the smaller the calibration set, the bigger the spread of the coverage will be around its average. This observation hints to an important problem, as the marginal coverage of a UM is precisely the average coverage over the ensemble of compatible calibration set samplings $\mathbb{E}(C)$, and is the main motivation for the technical contribution of this work, which we advance is presented in \autoref{sec:new_guarantee}.

\medskip \noindent 
Finally, in a real scenario we will not have an infinite test set, as initially assumed. The finiteness of the test set will introduce an extra layer of randomness to the coverage measurement. This extra randomness can be taken into account, instead of providing a pointwise measurement of coverage, by providing a confidence interval for the coverage. See more details in \ref{app:cov_test}.







\subsection{Statistical guarantees for a simple UM}
\label{sec:simple_um_guarantee_dem}

Continuing with the simple UM we provided as an example in \autoref{sec:simple_um}, now that we know how the coverage behaves, we can now explain how to use that information to provide a statistical guarantee of the marginal coverage we will get with this UM.\\
Assuming that we are building the UM with a desired coverage of $C^\text{nom}$ on a calibration set of size $N_\text{cal}$, we would need to compute a sample quantile $\hat{Q}_q$ of quantile level $q=C^\text{nom}$, which would result in the coverage $C$ being a random variable. When measured on an infinite test set (i.e. with no bias induced by finite-size effects of the test set), such random variable has a distribution
\begin{equation}
    \label{eqn:cov_dist}
    C \sim \text{Beta}(m, N_\text{cal}-m+1)
\end{equation}

\noindent with $m = \ceil{N_\text{cal}q} = \ceil{N_\text{cal}C^\text{nom}}$.
We can now quantify the deviation from the actual marginal coverage and the desired, nominal one:
\begin{multline}
   \mathbb{E}(C) - C^\text{nom} = \frac{m}{N_\text{cal}+1} - C^\text{nom} = 
    \begin{cases}
        \displaystyle-\frac{C^\text{nom}}{N_\text{cal}+1}\quad &\text{if} \; N_\text{cal}C^\text{nom} \in \mathbb{N}\\[5ex]
        \displaystyle\frac{1 - \left\{N_\text{cal}C^\text{nom}\right\} - C^\text{nom}}{N_\text{cal}+1} &\text{otherwise}
    \end{cases}
\end{multline}

\noindent where $\left\{\cdot\right\}$ is the fractional part function, which satisfies 
\begin{gather}
    x = \floor{x} + \{x\}\\
    0\leq\{x\}<1
\end{gather}
\noindent with $x\in\mathbb{R}$. Considering that $C^\text{nom}$ is bounded in $0\leq C^\text{nom} \leq 1$, the previous difference is bounded in the interval
\begin{equation}
    \frac{-1}{N_\text{cal}+1} \leq \mathbb{E}(C)-C^\text{nom} \leq \frac{1}{N_\text{cal}+1}
\end{equation}
\noindent which, recalling that $\mathbb{E}(C) = \mathbb{P}(\mathbf{Y}\in\mathscr{S})$, is the statistical guarantee that was presented in \autoref{sec:simple_um}.

\section{Conformal Prediction}
\label{sec:conf_pred}

Now that we have introduced the core concepts of this paper (surrogate modelling, and UMs and their coverage), we will briefly describe the conformal prediction framework, a popular method to transform any UM to a UM with some statistical guarantee.

\subsection{What is conformal prediction? An introductory example}
\label{sec:conf_pred_example}
Conformal prediction is a framework to build statistically robust UMs from UM models that do not offer any statistical guarantee on their coverage. The process of transforming a UM into one with statistical guarantees is called model conformalisation.
To illustrate the basics of the conformal prediction framework, we will now conformalise the simple UM we described in \autoref{sec:simple_um}, which we advance can already be seen as a conformal predictor built on top of a trivial uncertainty model, as we will explain now.\\
Recall the set up: we have built a surrogate regression model $\mathscr{F}$ which has been trained on the training dataset.  This surrogate model, as mentioned in \autoref{sec:preliminaries}, will necessarily have some approximation error that we will need to characterise and quantify in order to use it reliably. Specifically, we shall build prediction intervals associated with each of the predictions of the surrogate. To keep things as simple as possible, let us intentionally propose a totally trivial UM: one that always assumes a perfect prediction and thus postulates a null prediction error. This trivial $\widetilde{\mathscr{U}}$ is such that $\widetilde{\mathscr{U}}(\mathbf{X}, \mathbf{\hat{Y}})=\mathbf{U}=\mathbf{0}$ for all points. Given this baseline error quantification, the prediction interval shrinks to a single point $\mathscr{S} = \left[ \hat{\mathbf{Y}}-\mathbf{U}, \hat{\mathbf{Y}}+\mathbf{U}\right]=\hat{\bf Y}$ and, consequently, any minimal source of error makes the coverage of this UM equal to zero. Now, given this trivially useless UM, we can ask ourselves: can we use the information from the calibration set to make a modification to the UM's prediction intervals so that the achieved coverage $C$ and the desired coverage $C^\text{nom}$ fulfil some relationship? Let us conformalise the UM.\\
As it turns out, by adding a correction to the output of the UM, we will be able to assure that its coverage (or more specifically, its marginal coverage) will fulfil some properties. We will proceed in a similar fashion as we did in \autoref{sec:simple_um} when building the simple UM we introduced in that section, albeit with some subtle differences:

\begin{enumerate}
    \item First, we compute the absolute difference between the absolute error of $\mathscr{F}$ and the estimation $\mathbf{U}$ for all points in the calibration set. In other words, we define $s_i = \left|\mathbf{Y}_i - \hat{\mathbf{Y}}_i\right| - \mathbf{U}$ to be the difference on the $i$-th calibration example, hence obtaining the set $\left\{s_i\right\}_{i=1}^{N_\text{cal}}$. Each $s_i$ can be interpreted as a {\it score} that evaluates how bad the error estimation done by $\widetilde{\mathscr{U}}$ is: the higher $s_i$ is, the poorer the estimation of the error $\mathbf{U}$ is. In our trivial case, we have $\bf U=0$, and thus the scores $\left\{s_i\right\}_{i=1}^{N_\text{cal}}$ reduce in this trivial case to the absolute error (or in general to the error metric of choice) of the surrogate model for each point in the calibration set, but note that the procedure holds in general for other (less extreme) UMs and in the general case the scores $\left\{s_i\right\}_{i=1}^{N_\text{cal}}$ do not characterise anymore the errors of the surrogate predictions, but rather the mismatch between these errors and their quantification $\bf U$ via the UM.
    \item Second, we compute the sample quantile $\hat{Q}_q$ at level $q=C^\text{nom}$ in the set of scores $\left\{s_i\right\}_{i=1}^{N_\text{cal}}$.
    \item Finally, we use $\hat{Q}_q$ as a small correction to $\mathbf{U}$, such that the final prediction interval will be $\mathscr{S} = \left[ \hat{\mathbf{Y}}-\mathbf{U}-\hat{Q}_q, \hat{\mathbf{Y}}+\mathbf{U}+\hat{Q}_q\right]$. In our trivial case where $\bf U=0$, the resulting prediction interval is $\mathscr{S} = \left[ \hat{\mathbf{Y}}-\hat{Q}_q, \hat{\mathbf{Y}}+\hat{Q}_q\right]$.
\end{enumerate}

By following this procedure, it can be proved (in a similar fashion to what was done in \cite{papadopoulos2002inductive}) that the difference between the actual marginal coverage and the desired marginal coverage will be bound by \autoref{eqn:guarantee_simple}. Interestingly, when $\mathbf{U}=\mathbf{0}$, the conformalised prediction intervals precisely coincide with the ones obtained in the simple UM presented in \autoref{sec:simple_um}. So in summary, that simple UM model presented in \autoref{sec:simple_um} could actually be seen as the conformalisation of a totally trivial UM that initially assumes no error for the prediction of the surrogate model.\\
On a technical note, one very small modification must be done to improve the statistical guarantee that this UM provides and arrive to the statistical guarantee conformal predictors provide. By introducing the small correction 
\begin{equation}
    q^* = \frac{\ceil{(N_\text{cal}+1) C^\text{nom}}}{N_\text{cal}}
    \label{eqn:cp_correction}
\end{equation}
\noindent to the quantile level $q$ we are using to take into account the finite size of the calibration set, we reach the well-known statistical guarantee that conformal predictors provide
\begin{equation}
    0 \leq \mathbb{E}(C) - C^\text{nom} \leq \frac{1}{N_\text{cal}+1}
    \label{eqn:conf_pred_guarantee}
\end{equation}
\noindent which offers the advantage of being conservative, i.e., the marginal coverage will be \emph{at least} the desired coverage. The procedure to obtain \autoref{eqn:conf_pred_guarantee} can be found in \ref{app:conf_pred_guarantee}. This way, by employing the corrected quantile level $q^*$, the simple UM presented in \autoref{sec:simple_um} can be seen as a conformal predictor built on top of a null UM.

\medskip \noindent 
The previous simple example encapsulates the philosophy behind conformal prediction, which can be summarized in the following steps \cite{angelopoulos2021gentle}:

\begin{enumerate}
    \item Choose an uncertainty metric and build an heuristic UM, i.e. $\bf{U}=\widetilde{\mathscr{U}}(\bf X, \hat{Y})$. Given this heuristic UM, we compute an empirical distribution of a score function $s(\mathbf{X}, \mathbf{Y}, \hat{\mathbf{Y}})$ that encodes information on how well the UM $\widetilde{\mathscr{U}}$ approximates the true uncertainty of the predictions of the surrogate model $\mathscr{F}$.
    \item Given the empirical distribution of scores, compute the sample quantile $\hat{Q}_{q^*}$ (note that we are using the corrected version of the quantile level).
    \item Use the computed sample quantile to define the prediction set. In this case, the sample quantile acts as a small correction applied to the output of $\widetilde{\mathscr{U}}$.
\end{enumerate}

Observe that the UM model $\widetilde{\mathscr{U}}$ we want to conformalise is totally general and only requires to encode some  heuristic notion of the uncertainty of the surrogate's predictions. For instance, it can be a null heuristic, i.e. giving ${\bf U}=\mathbf{0}$: that is akin to assume that we don't have a UM to conformalise in the first place, in this case the conformal prediction recipe above is used to build from scratch a (conformal) UM with statistical guarantee. Alternatively, $\bf U$ can be estimated as some sort of average error. While our rule of thumb here is to do so in the calibration set (as we need to characterise generalisation error), nothing prevents to estimate $\bf U$ directly from the training set (training error), since we later conformalise the UM in the calibration set. Finally, as discussed in the introduction, $\widetilde{\mathscr{U}}$ could itself be a neural network --typically trained in the calibration set, but not necessarily-- that outputs an estimation of the absolute error associated to each prediction of the original surrogate model $\mathscr{F}$. 
In a nutshell, any heuristic UM of the surrogate model can be conformalised, and in the absence of a UM, conformal prediction offers a recipe to build one. The resulting conformalised UM --called a conformal predictor-- will provide the statistical guarantee given by \autoref{eqn:conf_pred_guarantee}.

\medskip \noindent 
As a last comment, observe that some surrogate models $\mathscr{F}$ already encode some notion of uncertainty, as it happens when performing so-called quantile regression \cite{romano2019conformalized}.
In this case, the output of $\mathscr{F}$ is a pair of quantile estimates that form an interval (interpreted as a prediction interval, although without any statistical guarantee).
Another example where the output of the surrogate $\mathscr{F}$ can be seen as an heuristic of uncertainty is when $\mathscr{F}$ is a soft classifier. In this setting, the outputs of $\mathscr{F}$ can be seen as the approximate probabilities that an example belongs to a given class \cite{sadinle2019least}. 

\begin{figure}[!h]
    \centering
    \includegraphics[width=0.8\linewidth]{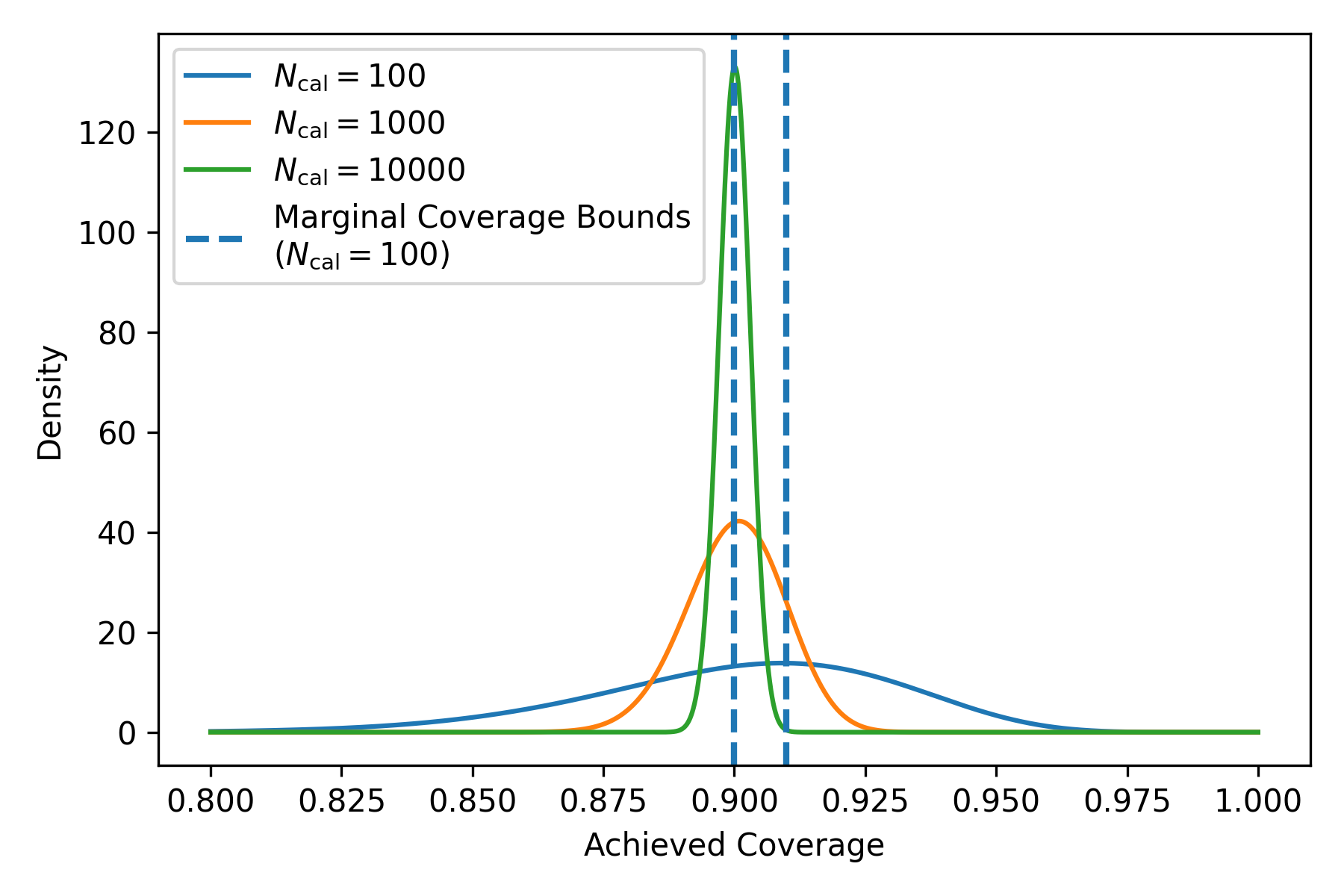}
    \caption{Coverage distribution (given by the pdf of a beta distribution) of a conformal predictor trained with different calibration set sizes, and a desired marginal coverage of $C^\text{nom}=0.9$. For the case of $N_\text{cal}=100$, the bounds given by \autoref{eqn:conf_pred_guarantee} are also displayed.}
    \label{fig:covdist}
\end{figure}

\subsection{What happens when $N_\text{cal}$ is small?}

The statistical guarantee of a conformal predictor (given by \autoref{eqn:conf_pred_guarantee}) is given in terms of bounds for the marginal coverage (i.e., the expected value of the coverage). However, as illustrated in \autoref{fig:covdist}, for sufficiently small calibration set sizes, the bounds provided by the guarantee are not representative of the actual coverage a conformal predictor will get: the spread of the coverage distribution for small values of $N_\text{cal}$ can cause the obtained coverage to be much smaller than the lower bound of the marginal coverage in a considerable proportion of cases. This is specially detrimental in cases where achieving satisfactory coverages for a single UM is key (as it happens in safety-critical applications), and where data is limited or costly to acquire, rendering small calibration set sizes. 

Importantly, observe that even if the original dataset is large, it is often the case that we need to conformalise the uncertainty model for different regions of the input space $S_X$. This is called group-balanced conformal prediction, and it seeks to obtain a conformal predictor that fulfils \autoref{eqn:conf_pred_guarantee} for a set of groups of $\mathbf{X}$. For example, if $\mathbf{X}$ is continuous, we would bin $\mathbf{X}$ and, by performing group-balanced conformal prediction, we would fulfil \autoref{eqn:conf_pred_guarantee} in every bin independently. In contrast, in conformal prediction as described in this section, we would fulfil \autoref{eqn:conf_pred_guarantee} in $S_X$ (globally), but if we evaluate the coverage in a given region of $S_X$, the coverage might not fulfil the guarantee. This is sensible to do when prediction errors are heterogeneously distributed throughout the input space, and is often the case when building surrogate models of physical models that for instance show strong gradients over their spatial domain. For all these reasons, 
being able to build UMs with relevant guarantees even for small calibration sets is certainly needed. In \autoref{sec:new_guarantee} we will propose a modification to conformal prediction to precisely achieve a statistical guarantee that is still relevant when the dataset size is small, thus producing conformal predictors that provide useful prediction intervals regardless of the calibration set size they are built on.


\begin{figure}[p]
    \centering
    \begin{subfigure}[b]{0.42\textwidth}
        \centering
        \includegraphics[width=\textwidth]{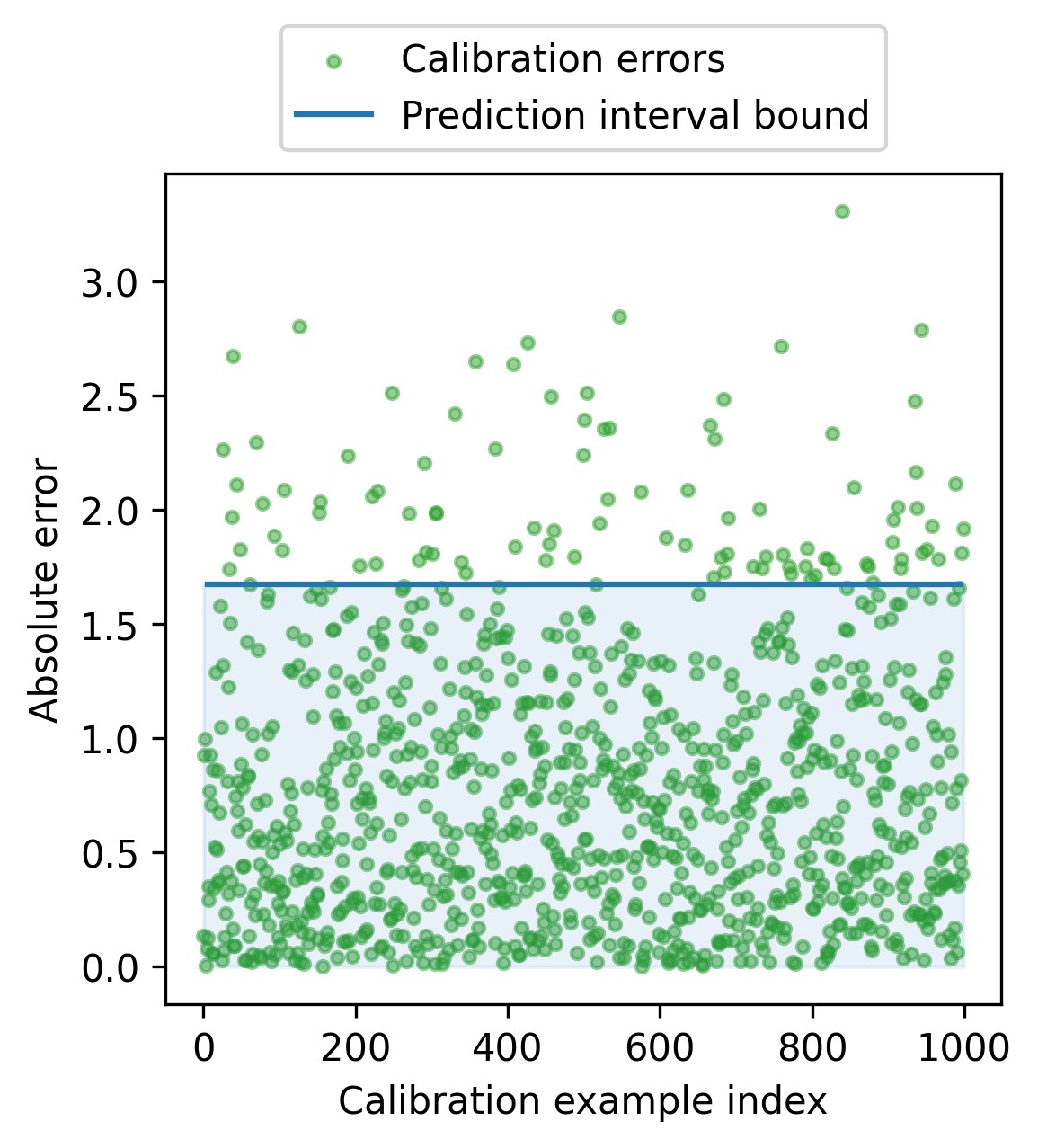}
        \caption{}
        \label{fig:simple_um_nlarge}
    \end{subfigure}
    \hfill
    \begin{subfigure}[b]{0.42\textwidth}
        \centering
        \includegraphics[width=\textwidth]{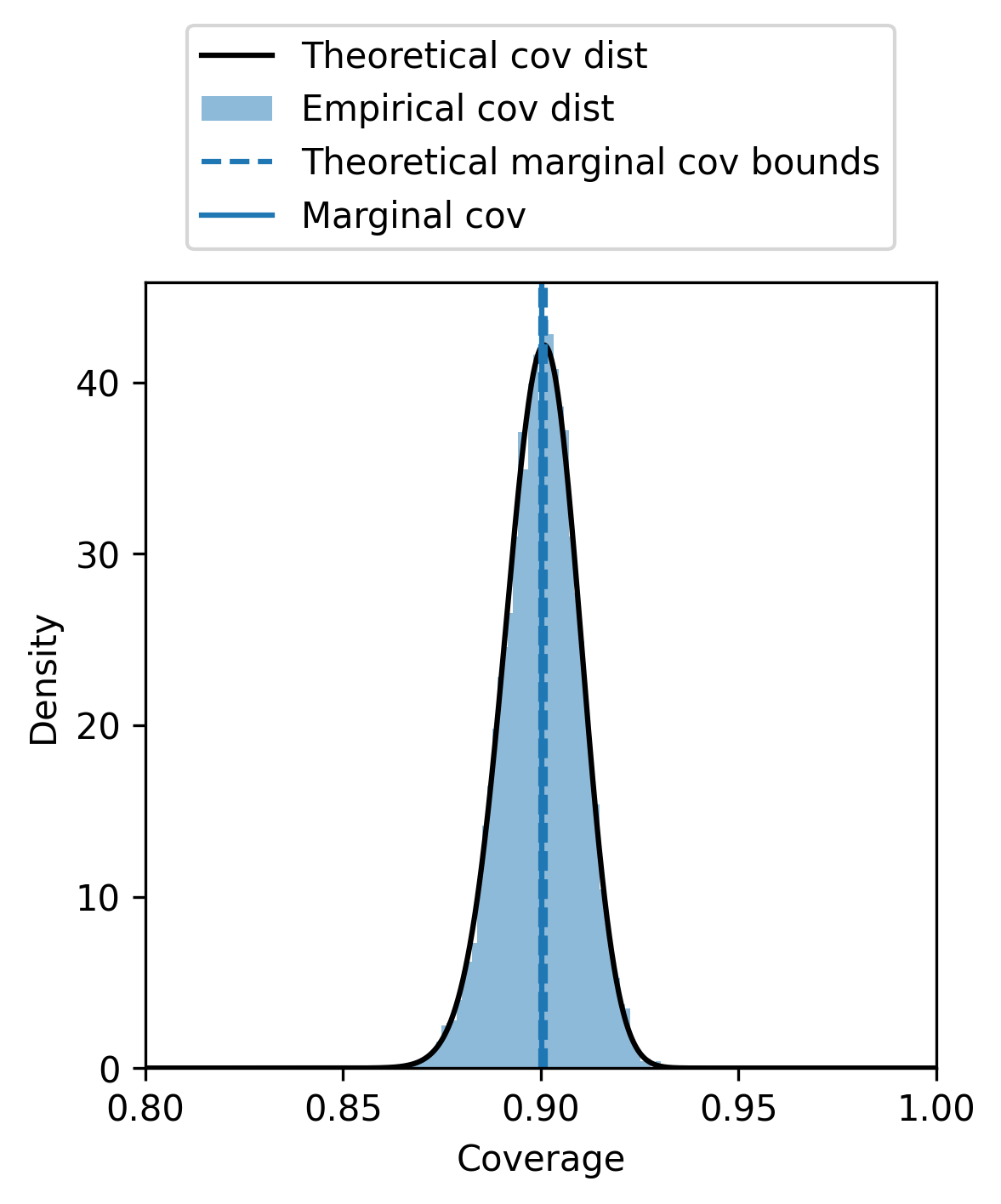}
        \caption{}
        \label{fig:cov_cpred_simple_nlarge}
    \end{subfigure}
    \\
    \begin{subfigure}[b]{0.42\textwidth}
        \centering
        \includegraphics[width=\textwidth]{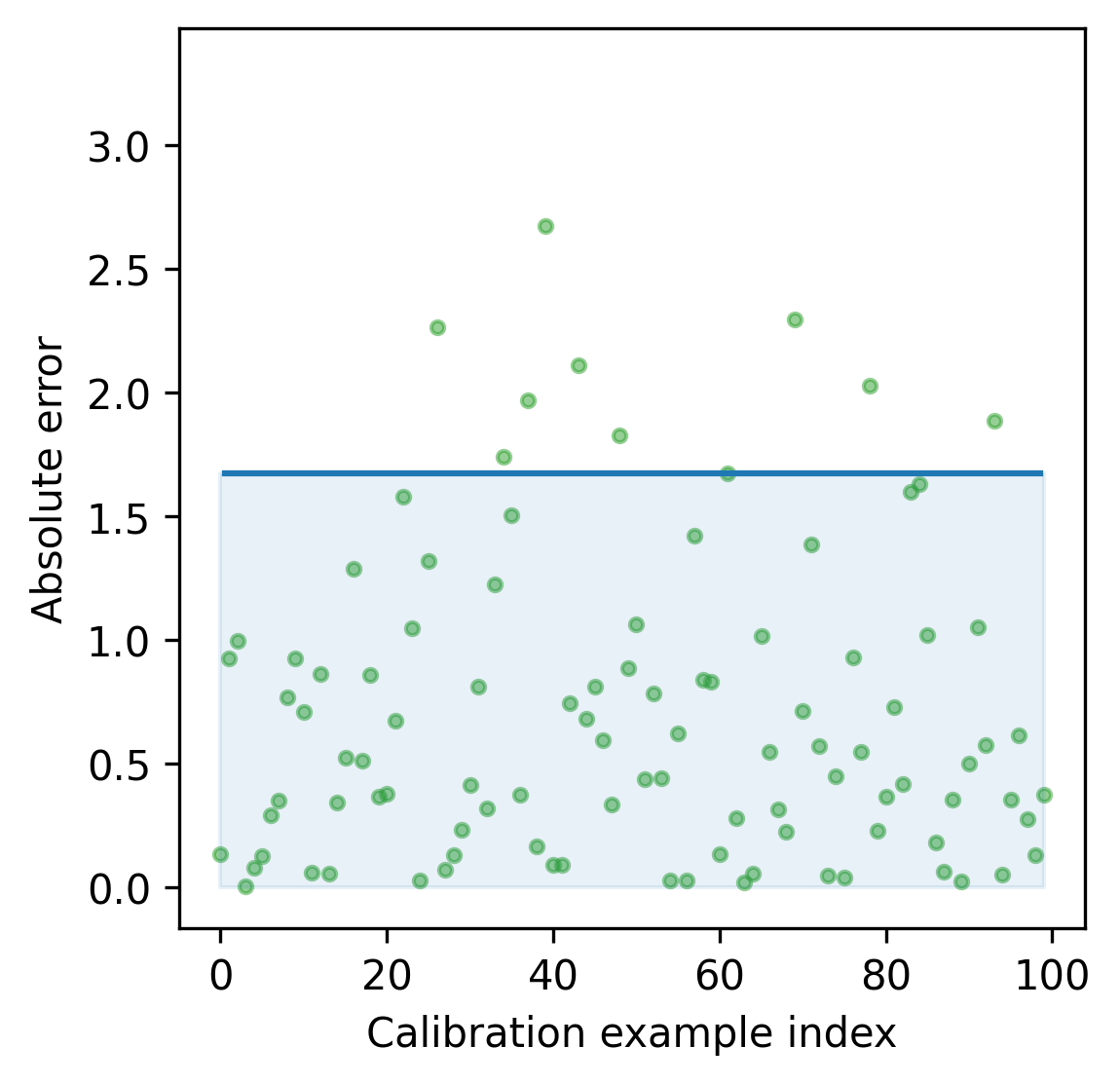}
        \caption{}
        \label{fig:simple_um_nsmall}
    \end{subfigure}
    \hfill
    \begin{subfigure}[b]{0.42\textwidth}
        \centering
        \includegraphics[width=\textwidth]{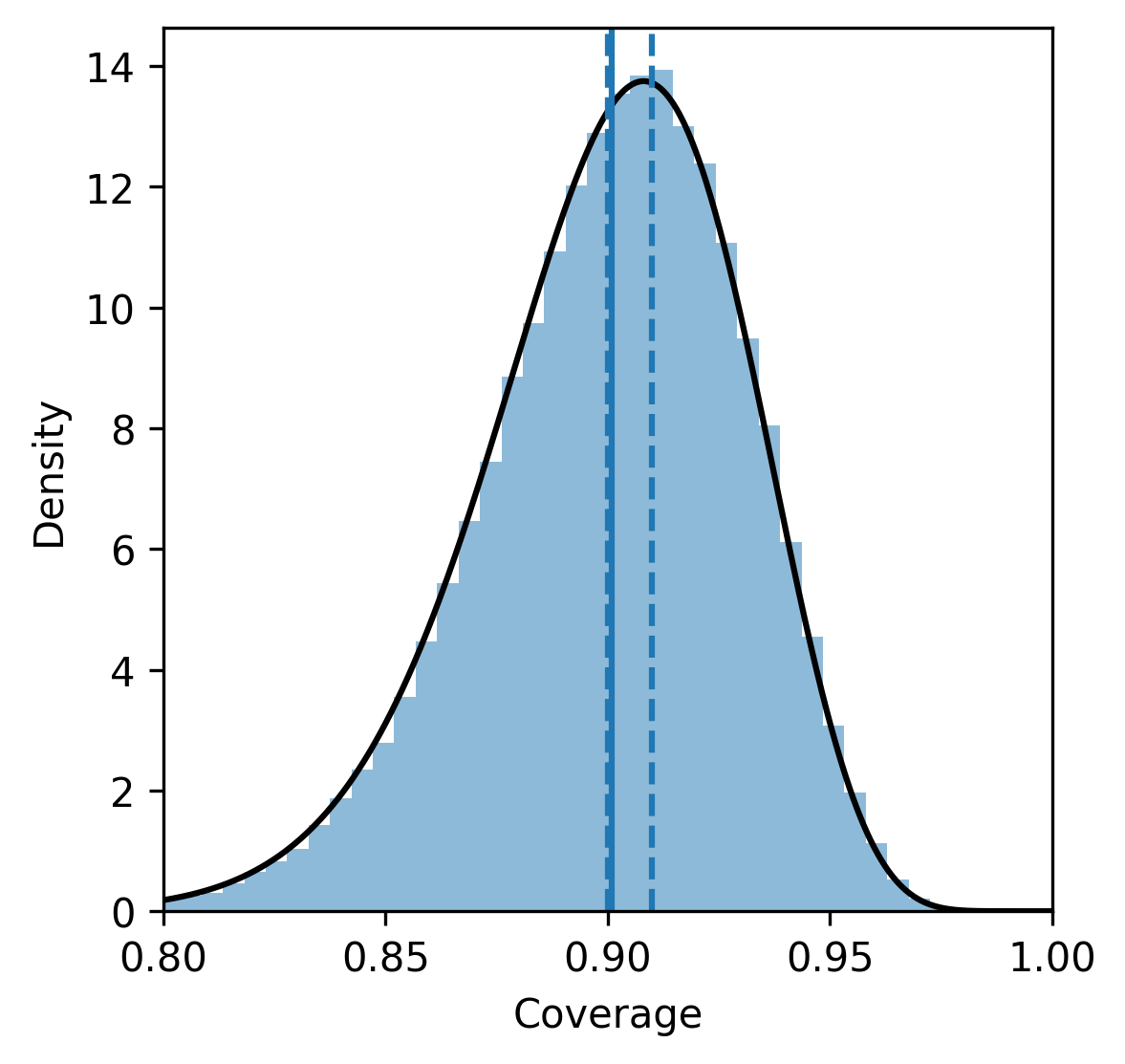}
        \caption{}
        \label{fig:cov_cpred_simple_nsmall}
    \end{subfigure}
    \caption{Illustrative example of the coverage distribution of a simple uncertainty model like the one described in \autoref{sec:conf_pred_example}, conformalising with a calibration set of size $N_\text{cal}=1000$ (top plots) and $N_\text{cal}=100$ (bottom plots). Left plots show the prediction interval (area shaded in blue) resulting from applying conformal prediction to a null uncertainty model using the green dots as calibration data. This represents one Monte Carlo realization. Right plots show the coverage distribution (empirical and theoretical) and the marginal coverage bounds obtained with the conformal prediction guarantee. The empirical distribution has been obtained through a Monte Carlo simulation of $N^{MC}=10000$ realizations.}
    \label{fig:simple_example}
\end{figure}

\subsection{A simple example}
\label{sec:conf_pred_sim}

To illustrate the characteristics of the standard conformal prediction guarantee and its dependence with the size of calibration set, we will use the simple conformal predictor described in \autoref{sec:conf_pred_example} along with some synthetic data.\\
Let us assume that the absolute error of a given surrogate model $\mathscr{F}$ can be sampled from a folded standard normal distribution (the distribution of the absolute value of a variable sampled from the standard normal distribution). The non-conformalised UM is the trivial one which systematically outputs a null prediction error ${\bf U}=0$, and we use a calibration set of size $N_\text{cal}=1000$ to conformalise the UM. The obtained coverages will be measured via the (analytical) cumulative distribution function of the assumed error distribution of $\mathscr{F}$. The conformal predictor will be built with a $C^\text{nom} = 0.9$. In \autoref{fig:simple_um_nlarge} we can see one example of what the calibration data and the corresponding prediction interval look like.\\
The process of training and conformalising the UM, and measuring its coverage is repeated a number of $N^{\text{MC}}$ Monte Carlo realisations, each time using a different sample from the data distribution and obtaining a different measure of coverage. The $N^{\text{MC}}$ measurements of coverage provide the empirical coverage distribution of the UM. In \autoref{fig:cov_cpred_simple_nlarge} we show the empirical coverage distribution obtained following this procedure. We compare this empirical distribution with the pdf of the analytical distribution provided by \autoref{eqn:cov_dist}, where a match is clearly observed. In this case, since the calibration set size $N_\text{cal}=1000$ is big enough, the dispersion of the coverage is sufficiently low for the bounds for the marginal coverage provided by \autoref{eqn:conf_pred_guarantee} to be representative enough of the actual coverage the UM will obtain.\\
In figures \autoref{fig:simple_um_nsmall} and \autoref{fig:cov_cpred_simple_nsmall} we show the results for the same procedure repeated using a smaller calibration set of size $N_\text{cal}=100$. We can clearly see that the dispersion of the coverage distribution has considerably increased, thus making the bounds for the marginal coverage provided by the conformal prediction guarantee non-relevant as indicators of the actual coverage a conformalised UM will achieve.\\
In the next section we propose a modification to the corrected quantile level used to build conformal predictors, obtaining a statistical guarantee that still provides useful information with small datasets.

\section{Towards relevant statistical guarantees with small datasets}
\label{sec:new_guarantee}

As seen in the previous sections, the statistical guarantee provided by conformal predictors becomes non-relevant when the calibration set size is small. Now, we aim to find a new guarantee that (i) matches the standard one provided by conformal prediction for large datasets, and (ii) unlike the standard approach, still provides useful information about the coverage even with small datasets. To that aim, we define a {\it minimum acceptable coverage} $C^\text{min}$ and build conformal predictors whose coverage $C$ is at least $C^\text{min}$ with a confidence of (at least) a minimum level specified by the user. Formally, we are looking for UMs that satisfy:
\begin{equation}
    \label{eqn:new_guarantee}
    \mathbb{P}\left(C \geq C^{\text{min}}\right) \geq 1 - \alpha
\end{equation}
\noindent where $1-\alpha$ is the minimum confidence with which we will obtain coverages above the specified $C^\text{min}$.\\
The key difference with respect to the classic conformal prediction guarantee (\autoref{eqn:conf_pred_guarantee}) is that in the classic guarantee we specify bounds for the expected value of the coverage, i.e. the marginal coverage. However, as already noted, when the calibration set is small, even if the expected coverage still fulfils \autoref{eqn:conf_pred_guarantee}, the large dispersion of the coverage distribution can cause the coverage of a single conformal predictor to deviate substantially from these bounds, making the classic guarantee effectively irrelevant in small datasets. On the other hand, with the new guarantee, we are establishing the minimum coverage we will achieve (with a given confidence) \emph{independently} of the calibration set size, thus putting more emphasis on the relevance of the statistical guarantee of the UM.

\medskip \noindent
In practical terms, the classic conformal prediction guarantee works as follows: for a given calibration set of size $N_\text{cal}$, we fix a desired marginal coverage $C^\text{nom}$ and compute a corrected quantile level $q^*$ that makes the conformal predictor fulfil \autoref{eqn:conf_pred_guarantee}. In contrast, under the new guarantee we fix a minimum acceptable coverage $C^\text{min}$ and search for the corrected quantile level $\tilde{q}$ that makes the conformal predictor satisfy \autoref{eqn:new_guarantee}. This ensures that the resulting predictor will achieve a coverage above $C^\text{min}$ with a confidence of at least $1-\alpha$. To do this, we can make use of the analytical expression of the coverage distribution given by \autoref{eqn:cov_dist}, and look for the roots of the function
\begin{equation}
    g(m) = F_C(C^{\text{min}};m) - (1-\alpha),
    \label{eqn:g_cnom}
\end{equation}
\noindent where $F_C(C^{\text{min}};m)$ is the cumulative distribution function of the coverage evaluated at $C^\text{min}$, and which depends on the parameter $m$ that defines the order statistic (see \autoref{sec:sample_quantile_estimators}) being used to build the conformal predictor, i.e. $m$ defines the corrected quantile level $q^*$ used to build the conformal predictor. 
For a given $C^{\text{min}}$, the function $g$ describes the difference in confidence level that we would obtain using a corrected quantile level $q^*$ (defined by the variable $m$) instead of the desired confidence level of $1 - \alpha$. Thus, by solving for the roots of $g$, we are finding the parameter $m$ that provides the desired minimum acceptable coverage with a confidence of \emph{exactly} $1 - \alpha$. Since $m$ is an integer (we are using order statistics as sample quantile estimators), a root may not exist, and we may need to look for the lowest $m$ that provides \emph{at least} a confidence of $1 - \alpha$.

\medskip \noindent
Since the family of beta distributions is continuous on its parameters for the range of interest for this application ($m >0$), we can treat the variable to optimize as a real number for the root finding problem. That is, instead of looking for the roots of $g(m)$ with $m \in \mathbb{N}$, we will look for the roots of $g(\bar{m})$ with $\bar{m}\in\mathbb{R}^+$. This, together with the function $g(\bar{m})$ being monotonic, allows us to use any root finding algorithm available in the common optimisation packages. Once $\bar{m}$ is found, $m$ can be recovered as $m = \ceil{\bar{m}}$, where we have chosen the ceiling function (instead of flooring) to get the most conservative interval.

\medskip \noindent
In addition to the approach just described (given a calibration set of size $N_\text{cal}$, compute the corrected quantile level $\Tilde{q}$ so that a conformal predictor fulfils \autoref{eqn:new_guarantee} with a given $C^\text{min}$ and confidence of at least $1-\alpha$), in \ref{app:new_guarantee} we explore two alternative scenarios in which

\begin{itemize}
    \item We want to compute the $C^\text{min}$ (for a given confidence level) of an already trained conformal predictor with the classic guarantee of \autoref{eqn:conf_pred_guarantee}.
    \item We want to know what is the minimum calibration set size $N_\text{cal}$ for which we can obtain a conformal predictor built with a quantile level of $\Tilde{q}$ and for a given $C^\text{min}$.
\end{itemize}

A Python implementation of all these procedures can be found in an open-access GitHub repository \cite{repo}, with an easy to use interface for employing the new guarantee with common conformal prediction libraries, such as MAPIE \cite{taquet2022mapie}. Below we illustrate this new method in a suite of examples.

\begin{figure}[p]
    \centering
    \begin{subfigure}[b]{0.42\textwidth}
        \centering
        \includegraphics[width=\textwidth]{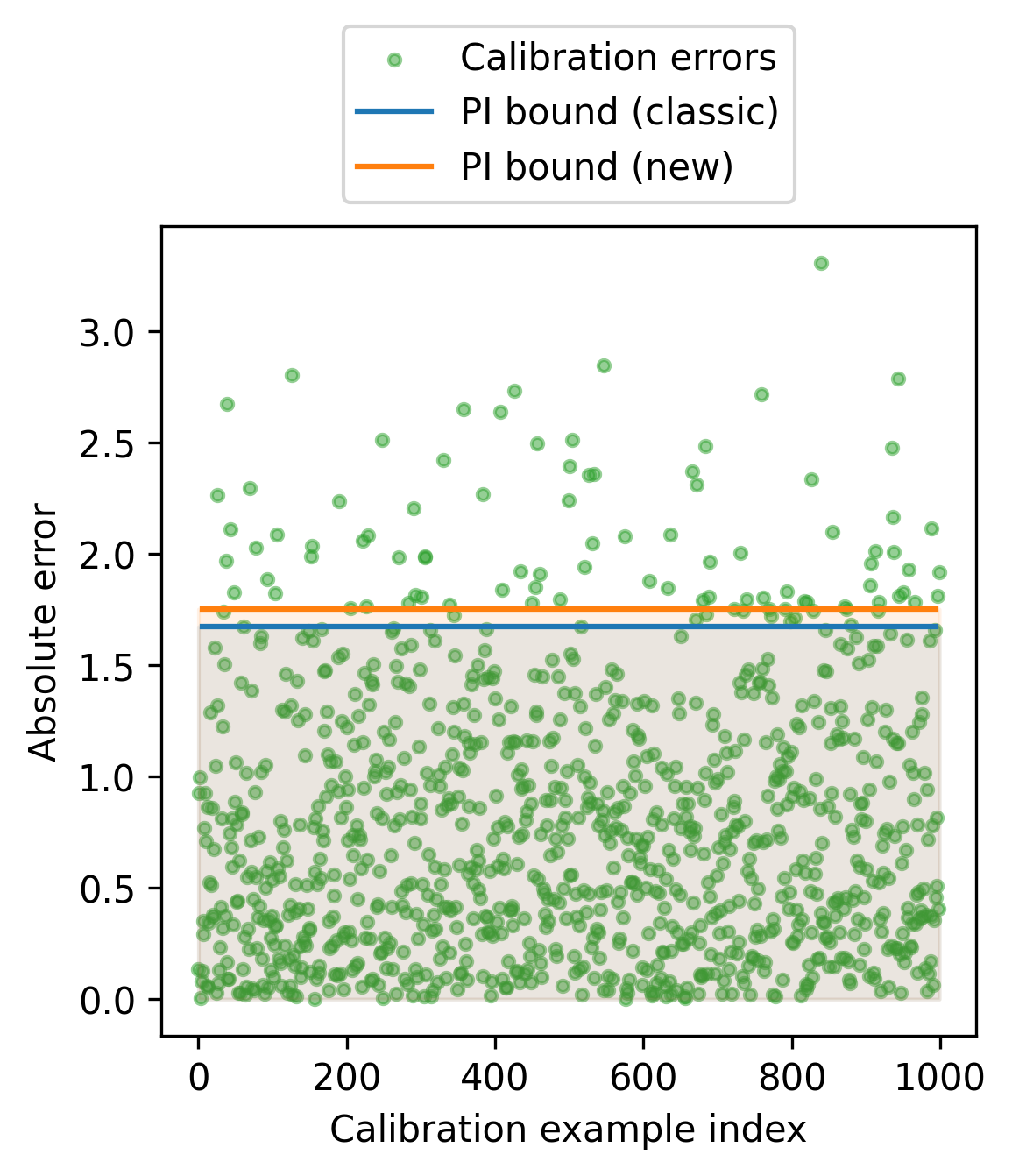}
        \caption{}
        \label{fig:simple_um_new_nlarge}
    \end{subfigure}
    \hfill
    \begin{subfigure}[b]{0.42\textwidth}
        \centering
        \includegraphics[width=\textwidth]{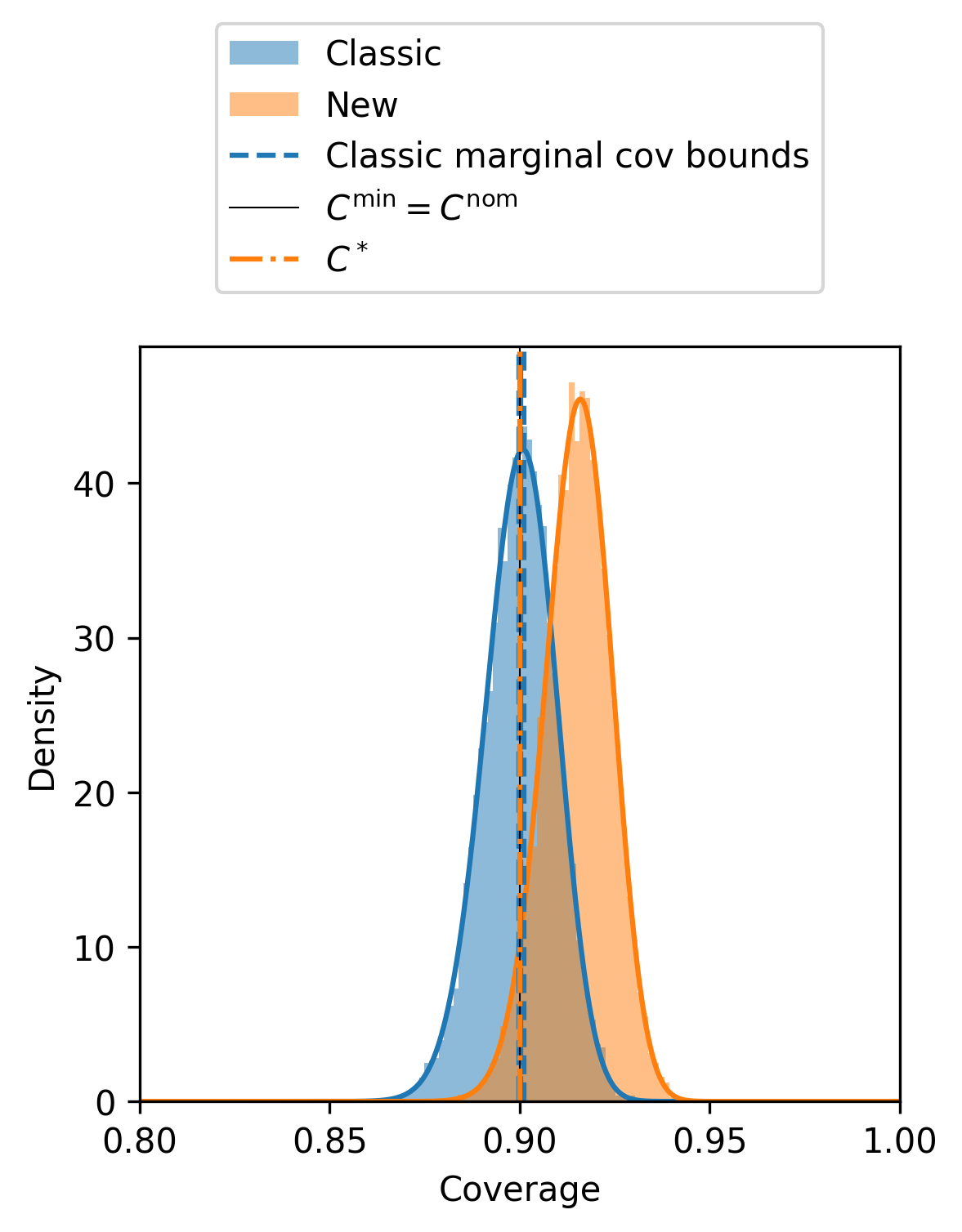}
        \caption{}
        \label{fig:cov_new_simple_nlarge}
    \end{subfigure}
    \\
    \centering
    \begin{subfigure}[b]{0.42\textwidth}
        \centering
        \includegraphics[width=\textwidth]{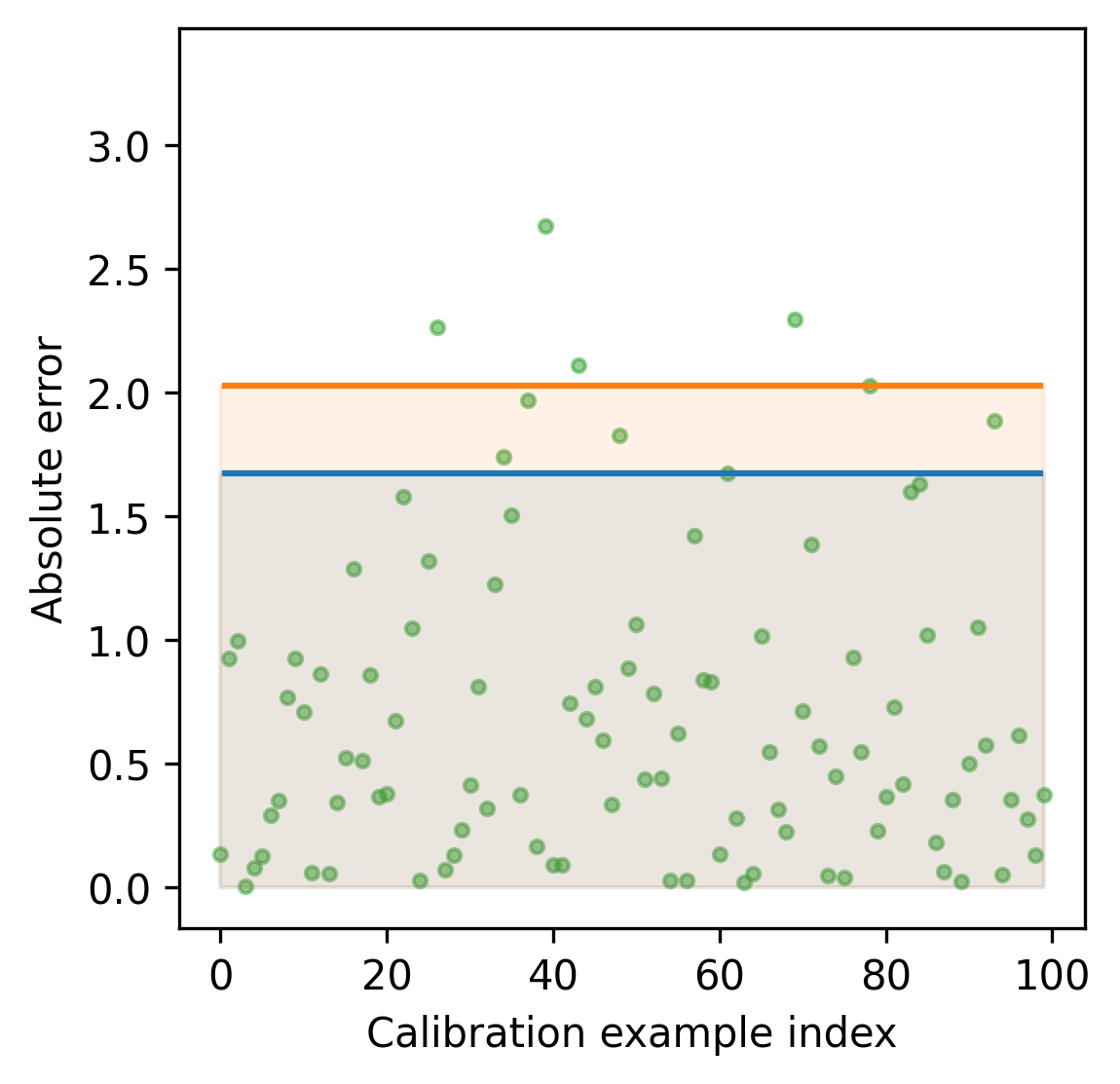}
        \caption{}
        \label{fig:simple_um_new_nsmall}
    \end{subfigure}
    \hfill
    \begin{subfigure}[b]{0.42\textwidth}
        \centering
        \includegraphics[width=\textwidth]{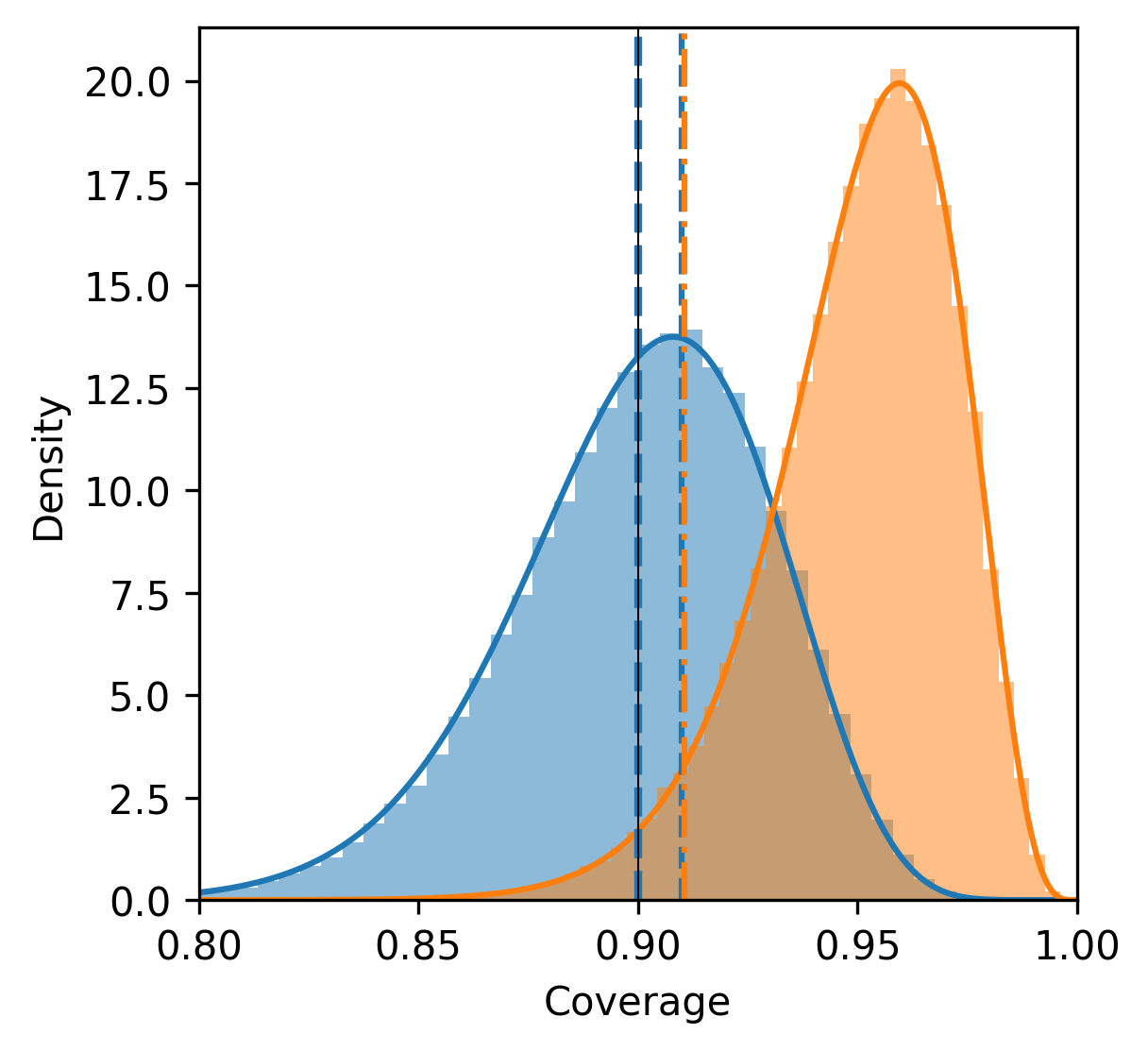}
        \caption{}
        \label{fig:cov_new_simple_nsmall}
    \end{subfigure}
    \caption{Comparison between the coverage distribution of a simple uncertainty model like the one described in \autoref{sec:conf_pred_example}, conformalized using the classic guarantee (\autoref{eqn:conf_pred_guarantee}, blue) and the new guarantee (\autoref{eqn:new_guarantee}, orange), using a confidence of $1-\alpha=0.95$ for the new guarantee and a calibration set of size $N_\text{cal}=1000$ (top plots) and $N_\text{cal}=100$ (bottom plots). Left plots show the prediction intervals (shaded areas) resulting from applying conformal prediction to a null uncertainty model using the green dots as calibration data. This represents one Monte Carlo realization. Right plot shows the coverage distribution and the marginal coverage bounds when using the classic conformal prediction guarantee (blue), and the coverage distribution with $C^\text{min}$ and $C^*$, i.e. the minimum coverage obtained for a confidence of $1-\alpha$ (orange). The empirical distribution has been obtained through a Monte Carlo simulation of $N^{MC}=10000$ realizations.}
    \label{fig:simple_example_new}
\end{figure}

\subsection{An illustrative example using a simple UM}
To illustrate the differences between a classical conformal with the guarantee \autoref{eqn:conf_pred_guarantee} and one using our proposed guarantee \autoref{eqn:new_guarantee}, we will continue to use the example of \autoref{sec:conf_pred_sim}.
In \autoref{fig:simple_example_new} we compare the coverage distribution of a conformal predictor built using the classic guarantee (blue) compared with the new guarantee (orange), for two calibration set sizes, $N_\text{cal}=1000$ (top plots of the figure) and $N_\text{cal}=100$ (bottom plots of the figure).
From these figures, it is clear that the new guarantee ``shifts'' the coverage distribution towards higher coverage values, enough so that we can ensure that \emph{any} conformal predictor will have a coverage above $C^\text{min}$ with at least a confidence of $1-\alpha$. This contrasts greatly with the coverage distribution provided by the classic guarantee, where a substantial proportion of the distribution is in a range of coverages lower than the desired $C^\text{nom}$, which can lead to conformal predictors with a coverage considerably lower than $C^\text{nom}$ when using small calibration sets. This is precisely what happens in the  $N_\text{cal}=100$ case. Observe in \autoref{fig:cov_new_simple_nsmall} that the spread of the coverage distribution is quite large for the classic guarantee. This makes the conformal predictor's statistical guarantee non-relevant: using \autoref{eqn:conf_pred_guarantee} we know that the \emph{marginal} coverage $\mathbb{E}(C)$ will be bounded in $\mathbb{E}(C) \in \left[C^\text{nom},C^\text{nom}+ 1/(N_\text{cal}+1)\right]\approx\left[0.9, 0.9099\right]$, but a considerable proportion of conformal predictors will end up having an actual coverage lower than 0.9. Specifically, around 46\% will have an actual coverage lower than 0.9, and around 10\% will have an actual coverage lower than 0.86. This loss of relevance of the classical guarantee is more acute when the calibration set is smaller, something that does not affect our newly proposed guarantee.

\subsection{On the informativeness of the new guarantee}
\begin{figure}[!h]
    \centering
    \includegraphics[width=0.49\linewidth]{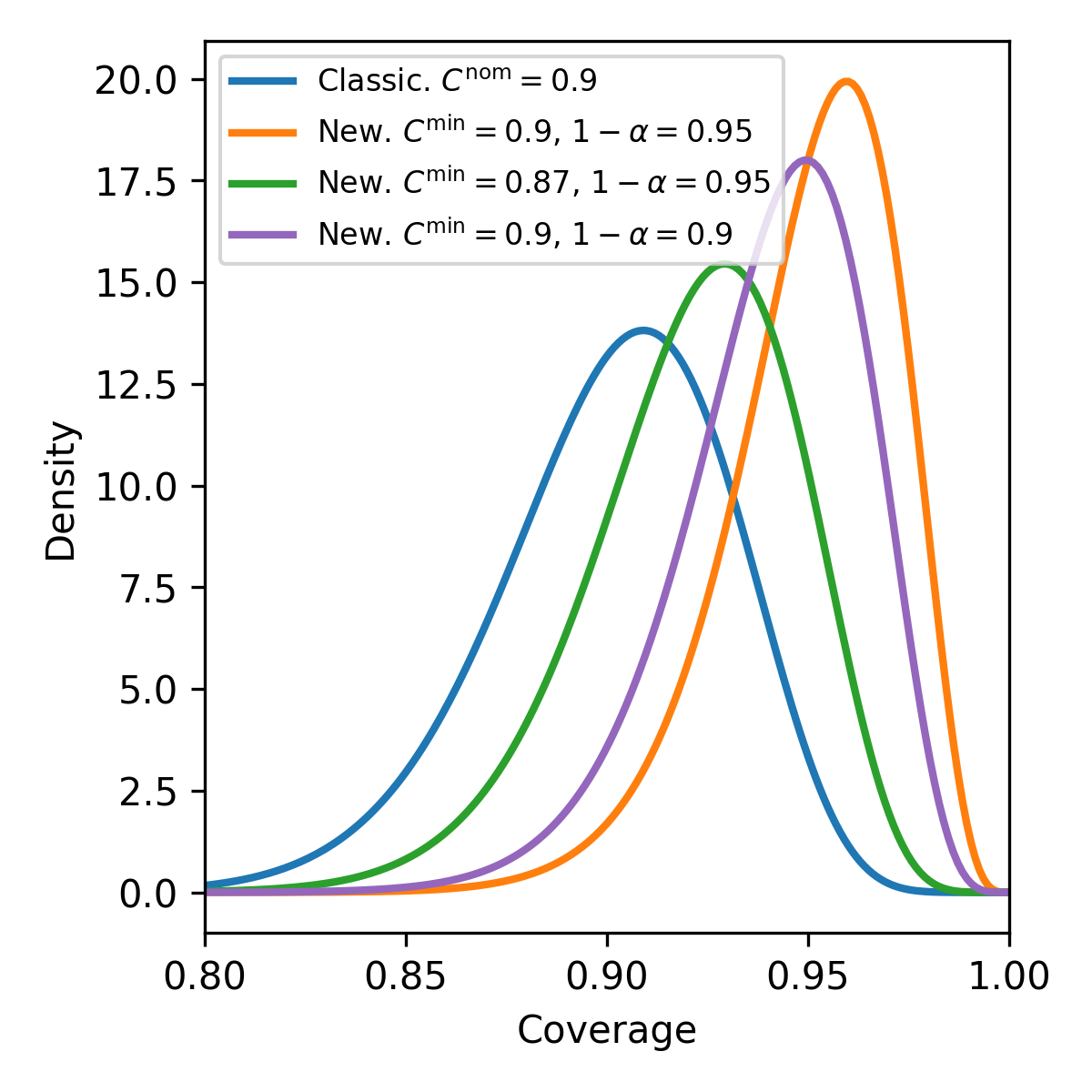}
    \caption{Coverage distributions given by a classic conformal predictor with $C^\text{nom}=0.9$ (blue), a conformal predictors with the new guarantee and varying $C^\text{min}$ and $\alpha$ (see legend).}
    \label{fig:cqr_ng_nicer}
\end{figure}

As seen in the previous section, by enforcing the new guarantee given by \autoref{eqn:new_guarantee} we achieve uncertainty models with relevant statistical guarantees even with small samples, where the classical conformalisation fails. However, these newly conformalised UMs are also more conservative than those obtained from classic conformal prediction, i.e. prediction intervals might be larger, potentially challenging the UM's informative character. 
While there is no way of knowing \emph{a priori} if this new type of conformal predictor will be sufficiently informative for a given minimum coverage and confidence (this must be evaluated a posteriori, e.g. by looking at the width of the prediction intervals in a regression problem), this potential loss of informativeness can be easily removed by lowering either the minimum acceptable coverage $C^\text{min}$ or the confidence $1-\alpha$. In \autoref{fig:cqr_ng_nicer} we present the coverage distributions of two alternatives to the UMs built so far with the new guarantee: one with a lower $C^\text{min}$, and another one with a lower confidence. Both options still fulfil \autoref{eqn:new_guarantee}, so they satisfy a relevant guarantee with small calibration sets, and offer slightly less conservative predictions than the UM presented in figures \ref{fig:simple_um_new_nsmall} and \ref{fig:cov_new_simple_nsmall}. Evidently, these relaxations in either $C^\text{min}$ or confidence must be carefully assessed in each particular use case.

\begin{figure}[!h]
    \centering
    \includegraphics[width=0.9\linewidth]{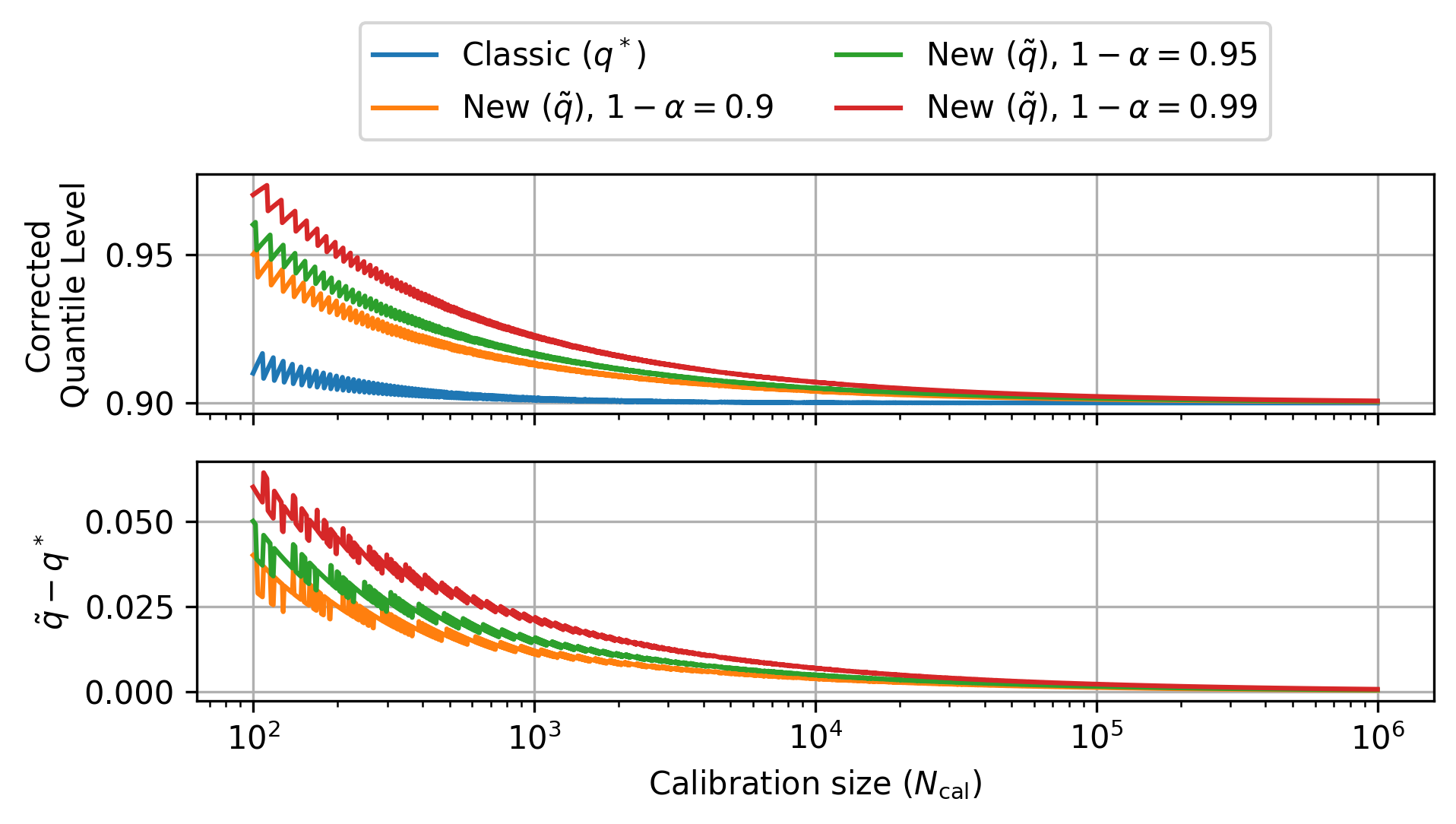}
    \caption{Comparison between the quantile level corrections obtained for the classic guarantee ($q^*$) and for the new guarantee ($\tilde{q}$) for different values of $1-\alpha$ and against the calibration sample size $N_\text{cal}$. These results are for the specifc case where $C^\text{nom}=C^\text{min}=0.9$.}
    \label{fig:old_vs_new_guarantee}
\end{figure}

\medskip \noindent
Finally, note that for large calibration set sizes, the classic and new guarantees converge to the same conformal predictor when $C^\text{min} = C^\text{nom}$. This is shown in \autoref{fig:old_vs_new_guarantee}, where we plot the corrected quantile level used to build a conformal predictor with the classic guarantee ($q^*$) and with the new one ($\tilde{q}$), against the calibration set size $N_\text{cal}$, for different values of $1-\alpha$. 



\section{Discussion}
\label{sec:discussion}

In this work, we have firstly provided a gentle introduction --almost in tutorial form-- to the relationship between surrogate models and their uncertainty models, as well as on the conformal prediction framework and how to conformalise uncertainty models. We have highlighted that standard conformal prediction yields non-relevant guarantees when the calibration set sizes are small, and have proposed a new set of guarantees to conformalise uncertainty models that are relevant independently of the calibration set size and therefore are still useful in those cases. These novel statistical guarantees --alongside a method to build conformal predictors with these guarantees-- represent the main technical contribution of this work.\\ 
Concretely, the proposed guarantee ensures that a given conformal predictor will achieve a coverage of at least $C^\text{min}$ with at least a confidence of $1-\alpha$. This guarantee is based on analytical expressions so, provided that the hypotheses on which conformal prediction is based are fulfilled, the validity of the information the guarantee provides is ensured when adequately conformalising any arbitrary uncertainty model.
The new guarantee renders very similar results to the classic guarantee for large calibration sample sizes and, unlike the classic guarantee, it is also relevant with small sample sizes. This relevance comes at a risk of producing too conservative conformal predictors, but this can be managed by slightly reducing the requirements in terms of minimum acceptable coverage and/or confidence. Conformal predictors built using our new guarantee will place special emphasis on the relevance of the guarantees they provide, making them particularly useful and well-suited to build UMs when data is scarce and/or when achieving satisfactory coverages for a single UM is essential: these two ingredients are present in safety-critical applications \cite{CoDANN} and in surrogate models of physical processes. In particular, our proposed conformalisation gains much relevance within group-balanced conformal prediction, i.e. when we aim at creating a UM that yields different prediction intervals for different regions of the parameter space. In this case, our method enables us to conformalise such region-dependent UM  and obtain a suite of independent conformal predictors conditioned to different regions, where all of them remain useful even if the region-dependent calibration sets are small.\\
In addition to our theoretical developments, we have implemented an open-access Python software implementation of this method with a user-friendly interface, which can be found
in a GitHub repository \cite{repo}.\\


\noindent{\bf Acknowledgments -- } 
The authors acknowledge funding from project TIFON (PLEC2023-010251) funded by MCIN/AEI/10.13039/501100011033, Spain.
LL acknowledges partial support from project CSxAI (PID2024-157526NB-I00) funded by \\ MICIU/AEI/10.13039/501100011033/FEDER, UE,
from a Maria de Maeztu Seal of Excellence grant (CEX2021-001164-M) funded by the MICIU/AEI/10.13039/501100011033, and from the European Commission Chips Joint Undertaking project No. 101194363 (NEHIL).\\

\noindent{\bf Data $\&$ Code --} Upon publication, our software solution will be available at \cite{repo}.


\appendix

\section{Derivation of the distribution of coverage}
\label{app:cov_dist}

As we have seen in \autoref{sec:cov}, if we use order statistics as sample quantile estimators (as in \autoref{eqn:samp_quantile}), the coverage follows a beta distribution. To arrive at such expression, first, we will consider that we have the set of absolute errors $\left\{s_i\right\}_{i=1}^{N_\text{cal}}$ resulting from evaluating the surrogate $\mathscr{F}$ on the calibration set. If we sort such errors we obtain

$$s_{1:N_\text{cal}} \leq s_{2:N_\text{cal}} \leq \dots \leq s_{i:N_\text{cal}} \leq \dots \leq s_{N_\text{cal}-1:N_\text{cal}} \leq s_{N_\text{cal}:N_\text{cal}}$$

\noindent being $s_{i:N_\text{cal}}$ the $i$-th order statistic, i.e. the $i$-th smallest value. Assuming that the errors follow an underlying probability distribution $\mathcal{D}$ (i.e. $s\sim \mathcal{D}$), the computed sample quantile $\hat{Q}_q$ will be an approximation of the true distribution quantile $Q_q$, at probability level $q$. As seen in \autoref{sec:cov}, this approximation will be at a probability level $\hat{q}$, which will be different from $q$.\\
If we assume that $\mathcal{D}$ is continuous, we can use the probability integral transform \cite{david1948probability} to express $F_s(s)$ in terms of a standard uniform distribution:
\begin{equation}
    F_s(s) \sim \text{Uniform}(0,1),
\end{equation}
then we can express the probability levels of the computed sample quantiles as order statistics of a random sample of size $N_\text{cal}$, iid of the standard uniform distribution. And, as it is well known \cite{gupta2004handbook}, these values follow the beta distribution:

\begin{equation}
    \label{eqn:sample_quantile_dist}
    \hat{q} = F_s(\hat{Q}_q) = F_s(s_{m:N_\text{cal}})= U_{m:N_\text{cal}} \sim \text{Beta}(m, N_\text{cal}-m+1)
\end{equation}

\noindent being $U_{m:N_\text{cal}}$ the $m$-th order statistic of a sample of size $N_\text{cal}$, iid from the standard uniform distribution.\\
The quantile level $\hat{q}$ provides the empirical quantile $\hat{Q}_q$ used to build the prediction interval  of the UM. Since the coverage is, by definition, the proportion of test examples inside the prediction interval, when computing the coverage on an infinite test set the coverage will be $\hat{q}$, thus being its distribution

\begin{equation}
    C \sim \text{Beta}(m, N_\text{cal}-m+1)
\end{equation}

\section{How do the calibration and test set sizes affect the coverage?}
\label{app:cov_test}

Uncertainty models here described are built on the hypothesis that the data used for their construction come from the same distribution as the data the UM will find in inference. In that sense, if the hypothesis perfectly holds and we could count on a sufficiently large test set to evaluate our UM, we would find that the coverage distribution in that hypothetical test set matches exactly the theoretical coverage distribution derived from the beta distribution.\\
However, in real world applications, the hypothesis of test and calibration sets coming from a single, common distribution will not hold as well as we would hope, so checking the coverage of a UM in a test set of unseen data is a good practice. And evidently, this test set will be finite. This finiteness of the test set will increase the variance in the estimation of the coverage. If we want to check the coverage of our UM, we need to account for this extra variance.\\
One way to do this is by representing the estimated coverage with its confidence interval. The probability of a new, previously unseen sample to fall inside the prediction interval of the UM will be its \emph{true} coverage $C$, i.e., the coverage that you would get on a infinitely large test set. This defines a Bernoulli process with a success probability of $C$. Thus, finding a confidence interval for $C$ translates into finding the binomial proportion confidence interval of such Bernoulli process. For that purpose, we can get an estimation for $C$ ---which we will use for building its confidence interval--- by measuring the coverage on the test set of size $N_\text{test}$, which we will denote by $\hat{C}$.\\
Since the coverages of interest are close to one (e.g. 0.95, 0.99, etc.), the computation of its confidence interval cannot rely on the usual approximation via the binomial distribution, thus the Clopper-Pearson interval \cite{clopper1934use} is preferred.

\section{Derivation of the conformal prediction guarantee}
\label{app:conf_pred_guarantee}

In a similar fashion to what we have done in \autoref{sec:simple_um_guarantee_dem}, here we derive the expression \autoref{eqn:conf_pred_guarantee} that provides the statistical guarantee that conformal predictors fulfil.\\
We will assume that we are building the conformal predictor with a desired coverage of $C^\text{nom}$ on a calibration set of size $N_\text{cal}$, but, in contrast to what we have done in \autoref{eqn:conf_pred_guarantee}, this time we will compute the sample quantile $\hat{Q}_{q^*}$ at the corrected quantile level 
\begin{equation}
    q^*=\frac{\ceil{C^\text{nom}(N_\text{cal}+1)}}{N_\text{cal}}
\end{equation}
\noindent which would result in the following coverage distribution (when measured on an infinite test set)
\begin{equation}
    \label{eqn:cov_dist_app}
    C \sim \text{Beta}(m, N_\text{cal}-m+1)
\end{equation}
\noindent with $m = \ceil{N_\text{cal}q^*} = \ceil{C^\text{nom}(N_\text{cal}+1)}$.\\
Once the conformal predictor has been built, the expected difference between its actual coverage and the desired $C^\text{nom}$ will be

\begin{multline}
    \mathbb{E}(C) - C^\text{nom} = \frac{m}{N_\text{cal}+1} - C^\text{nom} = 
    \begin{cases}
        \displaystyle0,\quad &\text{if} \; (N_\text{cal}+1)C^\text{nom} \in \mathbb{N}\\[3ex]
        \displaystyle\frac{1}{N_\text{cal}+1}\left(1 - \left\{(N_\text{cal}+1)C^\text{nom}\right\}\right), &\text{otherwise}
    \end{cases}
\end{multline}

\noindent where $\left\{\cdot\right\}$ is the fractional part function. Considering that $C^\text{nom}$ is bounded in $0\leq C^\text{nom} \leq 1$, the previous difference is bounded in the interval

\begin{equation}
    0 \leq \mathbb{E}(C)-C^\text{nom} \leq \frac{1}{N_\text{cal}+1}
\end{equation}

\noindent which is the statistical guarantee conformal predictors provide, as said in \autoref{sec:conf_pred}.

\section{Other procedures to achieve the guarantee given by \autoref{eqn:new_guarantee}}
\label{app:new_guarantee}

In \autoref{sec:new_guarantee}, we detailed how to build conformal predictors that satisfy the guarantee given by \autoref{eqn:new_guarantee} with a given dataset. Additionally, in this appendix, we explore another two methods to make a conformal predictor comply with \autoref{eqn:new_guarantee}.

\subsection{Compute $C^{\text{min}}$ given $N_\text{cal}$, $C^{\text{nom}}$ and $\alpha$}
\label{sec:proc_cmin}
If we have an already built conformal predictor that uses the classic quantile correction given by \autoref{eqn:cp_correction}, we can easily compute the minimum coverage we will get with a confidence of $1-\alpha$. Assuming we trained the conformal predictor on a calibration set of size $N_\text{cal}$, $C^{\text{min}}$ will be given by:
\begin{equation}
    C^{\text{min}} = F_C^{-1}(\alpha),
    \label{eqn:c_min}
\end{equation}
\noindent where $F_C^{-1}$ represents the inverse of the cumulative distribution function of the coverage distribution, i.e. the percent point function. Since the coverage distribution follows the distribution given by \autoref{eqn:cov_dist_app}, the numerical value of \autoref{eqn:c_min} can be easily computed, since $F_C^{-1}$ (the percent point function of a beta distribution) is implemented in most statistical software packages.

\subsection{Compute $N_\text{cal}$ min given $C^{\text{min}}$, $C^{\text{nom}}$ and $\alpha$}
\label{sec:proc_nmin}

\begin{figure}[!h]
    \centering
    \includegraphics[width=0.9\linewidth]{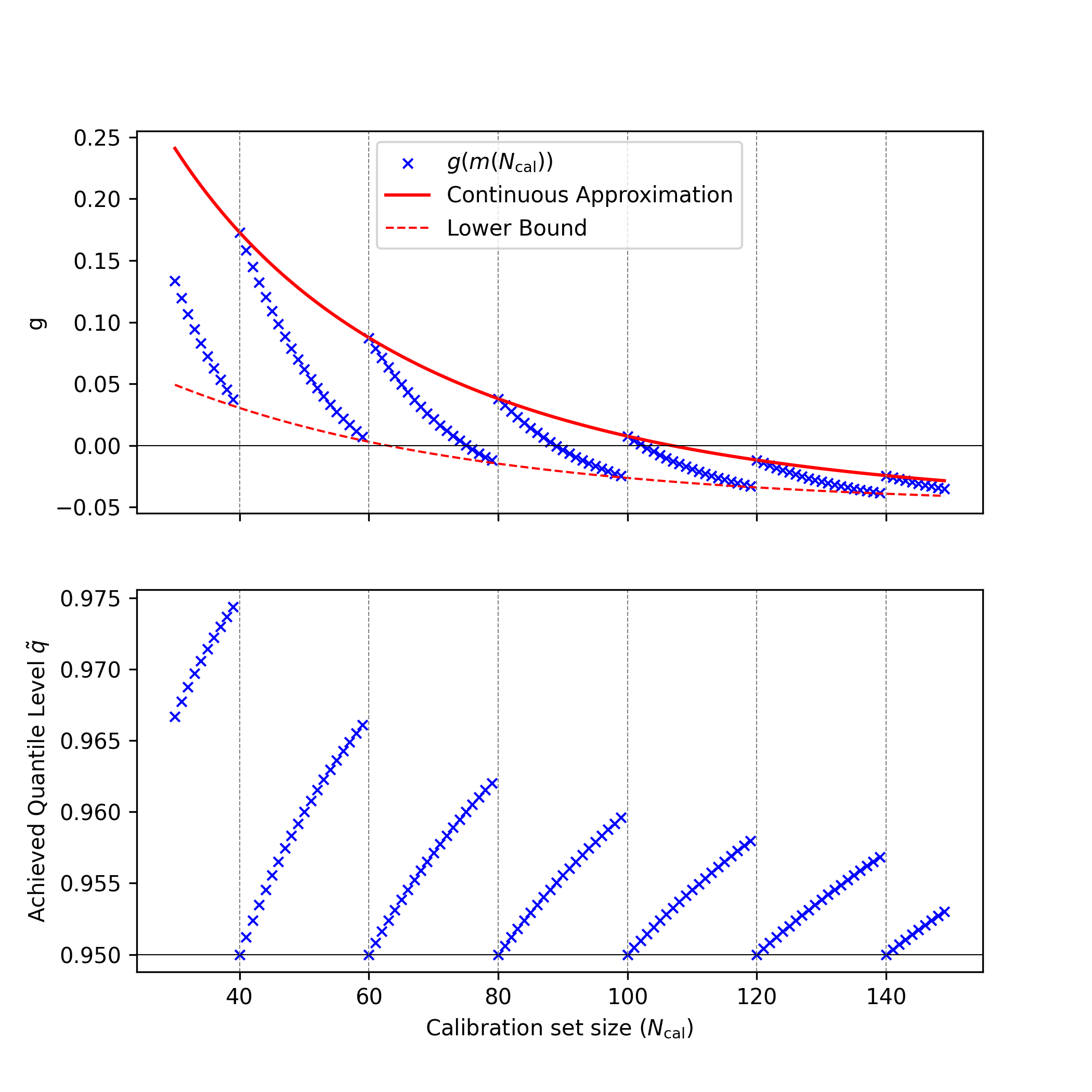}
    \caption{Search for the minimum $N_\text{cal}$ that allows a conformal predictor with a minimum coverage of $C^{\text{min}} = 0.9$ with a confidence of $1-\alpha=0.95$, and built with a quantile level of $\tilde{q} \approx 0.95$. Top plot displays the function $g(m(N_\text{cal}))$, and bottom plot the achieved quantile level. Blue crosses represent actual models ($m \in \mathbb{N}$), the red continuous line shows the approximation considering $\bar{m}$ (i.e. the upper bound of $g(m(N_\text{cal}))$, see text for details), and the red discontinuous line represents the lower bound of $g(m(N_\text{cal}))$. With thin, vertical lines we highlight values of $N_\text{cal}$ for which $m=\bar{m}$, thus achieving the quantile level \emph{exactly}.}
    \label{fig:n_search_beta}
\end{figure}

If we are in the stage of acquiring new data, we might want to get enough examples to build a conformal predictor built with quantile level $\tilde{q}$, and that satisfies \autoref{eqn:new_guarantee} for a given $C^{\text{min}}$. In other words, we might wonder: what is the minimum calibration set size for building a conformal predictor with a quantile level of $\tilde{q}$, and that reaches at least $C^{\text{min}}$ with a confidence of $1-\alpha$? In order to answer to this question, a process very similar to the described in \autoref{sec:new_guarantee} suffices. We will look for roots of the function
\begin{equation}
    g(m(N_\text{cal})) = F_C(C^{\text{min}};m(N_\text{cal})) - (1-\alpha)
\end{equation}
\noindent where now the parameter $m$ depends on the, now variable, calibration set size $N_\text{cal}$.\\
Since we are working with order statistics as sample quantiles, asking to match \emph{exactly} the desired quantile level might be too restrictive and prevent the existence of a root. To avoid this, we allow some slack to the quantile level, looking for solutions that provide the minimum coverage for the desired confidence and a quantile level that is \emph{close} to the desired one. This is precisely what we show in the bottom plot of \autoref{fig:n_search_beta}, where we display the achieved quantile level $\tilde{q}$ for a given minimum coverage. In the top plot of the same figure, with blue crosses, we display the value of ${g(m(N_\text{cal}))}$.\\
Additionally, to ease the root-searching process, instead of $m$, we can use a continuous version of this parameter
\begin{gather}
    \bar{m}=N_\text{cal}\tilde{q}.
\end{gather}
\begin{sloppypar}   
This continuous approximation renders the red continuous line of the top plot in \autoref{fig:n_search_beta} which, as it can be seen, is the upper bound for the function ${g(m(N_\text{cal}))}$. Moreover, we can also find a lower bound for ${g(m(N_\text{cal}))}$ as a function of $\bar{m}$, which is given by considering the distribution ${C^\text{inf}\sim\text{Beta}(\bar{m}+1, N_\text{cal}-\bar{m})}$ (see \ref{app:nmin_bounds} for a complete justification for these bounds).
\end{sloppypar}
The roots of these two functions ---the upper and lower bounds of $g(m(N_\text{cal}))$---  give us two values of $N_\text{cal}$: $N_\text{cal}^\text{sup}$, and $N_\text{cal}^\text{inf}$.\\
The value $N_\text{cal}^\text{inf}$ indicates a lower bound for the wanted $N_\text{cal}$: we will not be able to achieve the desired minimum coverage with a confidence of $1-\alpha$ if the calibration set has a size of $N_\text{cal} < N_\text{cal}^\text{inf}$. Similarly, $N_\text{cal}^\text{sup}$ defines an upper bound for the calibration set size: any model built with $N_\text{cal} > N_\text{cal}^\text{sup}$ will achieve the minimum coverage with a confidence level higher than $1-\alpha$.

\subsection{Derivation of bounds for $g(m(N_\text{cal}))$}
\label{app:nmin_bounds}
We will focus now on finding bounds for $F_C$ in terms of CDFs of beta distributions with parameters $\bar{a} = \bar{m}$, $\bar{b} = n - \bar{m} + 1$. We will consider that $F_C$ was given by the CDF of a beta distribution $\text{Beta}(a, b)$, with $a = m$, $b = n - m + 1$. First, we begin by finding an expression for the difference $\bar{a} - a$
\begin{equation}
    \bar{a} - a = \bar{m} - m = \begin{cases}
    0,\quad &\text{if} \; N_\text{cal}p \in \mathbb{N}\\
    \{N_\text{cal}p\}-1,\quad &\text{otherwise}
    \end{cases}
\end{equation}
\noindent where $\{\cdot\}$ is the fractional part function which satisfies $x = \floor{x} + \{x\}$. Similarly, the difference $\bar{b} - b$ can be expressed as
\begin{equation}
    \bar{b} - b = (N_\text{cal} - \bar{m} + 1) - (N_\text{cal}-m+1) = \bar{m}  - m = \bar{a} - a.
\end{equation}
Since the fractional part of any real number $x$ satisfies $0 \leq \{x\} < 1$, both $\bar{a} - a$ and $\bar{b} - b$ can be bounded in the interval $(-1, 0]$. Taking into account that the CDF of $\text{Beta}(a, b)$ is monotonically increasing in $b$ and monotonically decreasing in $a$. We can finally bound $F_C(C^{\text{min}};m(N_\text{cal}))$ as
\begin{equation}
   F_{C^{\text{inf}}}(C^{\text{min}}; \bar{m}(N_\text{cal})) \leq F_C(C^{\text{min}}; m(N_\text{cal})) \leq F_{C^{\text{sup}}}(C^{\text{min}}; \bar{m}(N_\text{cal})),
\end{equation}
\noindent where $C^\text{inf} \sim \text{Beta}(\bar{a}+1, \bar{b}-1)$ and $C^{\text{sup}} = \text{Beta}(\bar{a}, \bar{b})$.


\bibliographystyle{elsarticle-num}
\bibliography{cas-refs}
\end{document}